\documentclass[pdftex,epjc3]{svjour3}          
\usepackage[
    a4paper,
    left=2.5cm,
    right=2.5cm,
    top=2.5cm,
    bottom=2.5cm
]{geometry}
\RequirePackage[T1]{fontenc}

\smartqed  

\RequirePackage{graphicx}
\RequirePackage{mathptmx}      
\RequirePackage{flushend}
\RequirePackage[numbers,sort&compress]{natbib}
\RequirePackage[colorlinks,citecolor=blue,urlcolor=blue,linkcolor=blue]{hyperref}

\journalname{Supercomputing}
\usepackage{graphicx}%
\usepackage{multirow}%
\usepackage{amsmath,amssymb,amsfonts}%
\usepackage{mathrsfs}%
\usepackage[title]{appendix}%
\usepackage{xcolor}%
\usepackage{textcomp}%
\usepackage{manyfoot}%
\usepackage{booktabs}%
\usepackage{algorithm}%
\usepackage{algorithmic}%
\usepackage{listings}%
\usepackage{tikz}
\usepackage{pgfplots}
\pgfplotsset{compat=1.18}
\usetikzlibrary{arrows.meta, calc, patterns, shapes, positioning}
\usepackage{amsmath}
\usepackage[c2]{optidef}
\usepackage{subcaption}
\usepackage{array}
\usepackage{placeins}

\begin{document}

\title{Exposure–Length Trade-off Optimization for Obstacle-Aware Intrusion Paths in Heterogeneous Wireless Sensor Networks}


\author{Nguyen Thi My Binh\thanksref{e1,addr1}
\and Ho Viet Duc Luong\thanksref{e2,addr2}
\and Nguyen Tuan Nghia\thanksref{e3,addr2}
\and Nguyen Quang Huy\thanksref{e4,addr2}
\and Dang Trong Hop\thanksref{c1,e5,addr1}
}
\thankstext[$\star$]{c1}{Corresponding author.}
\thankstext{e1}{e-mail: binhntm@haui.edu.vn}
\thankstext{e2}{e-mail: luonghvd@soict.hust.edu.vn}
\thankstext{e3}{e-mail: nghia.nt235179@sis.hust.edu.vn}
\thankstext{e4}{e-mail: luonghvd@soict.hust.edu.vn}
\thankstext{e5}{e-mail: hopdt@haui.edu.vn}

\institute{Hanoi University of Industry, 298 Cau Dien Street, Hanoi, 100000, Vietnam\label{addr1}
          \and
           Hanoi University of Science and Technology, 1 Dai Co Viet, Hanoi, 100000, Vietnam\label{addr2}
}

\date{Accepted for publication}

\maketitle

\begin{abstract}
Barrier coverage is a key paradigm for intrusion detection and surveillance in IoT-enabled heterogeneous wireless sensor networks, where monitoring quality depends not only on the existence of sensing barriers but also on the detectability of feasible crossing paths. In this paper, we study the obstacle-aware maximal exposure path planning problem in heterogeneous sensor fields under truncated omni-directional sensing and line-of-sight occlusion. Unlike conventional maximal exposure formulations that focus only on exposure maximization, the considered problem seeks side-to-side feasible paths that simultaneously maximize accumulated sensing exposure and minimize path length in obstacle-constrained environments. To address this bi-objective problem, we develop OG-MOEA/D, a problem-specific MOEAD method that combines a fixed-grid path representation with obstacle-aware feasibility checking and adaptive repair. The standard decomposition, neighborhood update, SBX, and polynomial mutation mechanisms are inherited from MOEA/D rather than claimed as new. Extensive simulations on networks with varying sensor densities and obstacle layouts show that OG-MOEA/D consistently improves hypervolume over generations and produces well-distributed Pareto fronts. Direct obstacle-constrained comparisons with MOEA/D and NSGA-II over 20 independent runs show higher HV and lower IGD for OG-MOEA/D in all four tested sensor-network sizes. The results should be interpreted as controlled simulation evidence for surveillance-strength evaluation in synthetic two-dimensional heterogeneous WSN scenarios, rather than as a model of realistic evasive intrusion behavior or a complete real-world deployment.
\end{abstract}

\section{Introduction}\label{sec:introduction}

Wireless sensor networks (WSNs) constitute a core sensing infrastructure for large-scale IoT-enabled cyber--physical systems due to their low cost, scalability, and flexible deployment characteristics \cite{gulati2022review,avila2025energy}. They have been widely adopted in security surveillance, border protection, environmental monitoring, industrial inspection, and disaster response, where distributed sensors collaboratively monitor a region of interest (ROI) and support situation awareness and decision-making \cite{oztoprak2024security,cacciuttolo2024internet}. In such applications, sensing performance depends not only on where sensors are deployed, but also on how effectively the network can detect traversal events across the monitored region.
Among the most important sensing-performance criteria, \emph{coverage} plays a central role in evaluating detection capability \cite{zhong2024hybrid,song2024wireless,anusuya2024comprehensive,moslehi2025exploring}. Within this context, \emph{barrier coverage} has attracted considerable attention because of its direct relevance to intrusion detection and border monitoring. Instead of requiring full-area sensing, barrier coverage aims to establish a virtual sensing boundary such that any side-to-side traversal across the ROI is detectable \cite{tossa2025wireless,binh2024efficient}. This paradigm is particularly suitable for belt-shaped or rectangular regions and provides an efficient sensing strategy when resources are limited \cite{maheshwari2019survey,egwuche2023machine,binh2024efficient}.

In practical IoT-enabled WSN deployments, however, the sensing environment is rarely homogeneous or obstacle-free. Real systems often consist of heterogeneous sensors with different sensing ranges, attenuation characteristics, sampling capabilities, and energy budgets \cite{kumar2022thwsn,kumar2025improved}. Such heterogeneity produces spatially non-uniform sensing fields, making both coverage assessment and traversal analysis substantially more difficult \cite{juwaied2025enhancing,binh2025efficient}. In addition, realistic ROIs frequently contain physical obstacles---such as buildings, lakes, vegetation, or restricted zones---that constrain feasible movement and deployment while also occluding sensing signals \cite{binh2022family}. These factors make obstacle-aware modeling essential and call for algorithmic methods that can jointly account for sensing heterogeneity, geometric feasibility, and line-of-sight (LoS) effects.
Beyond verifying whether a sensing barrier exists, an important complementary research direction focuses on explicitly identifying \emph{penetration paths}. In particular, the \emph{maximal exposure path} (MaEP) problem seeks a traversal path across the ROI that maximizes cumulative sensing exposure, thereby characterizing the network's best-case detection capability under constrained motion \cite{hong2017finding,wang2020generalized,van2023efficient}. {The maximal exposure path is therefore most naturally interpreted as a surveillance-strength or stress-testing construct: it identifies feasible crossings that remain strongly exposed. It should not be interpreted as the route of a rational evasive intruder; vulnerability-oriented analysis is more directly associated with minimal-exposure paths.} Nevertheless, MaEP is computationally challenging: even simplified variants are NP-hard \cite{lee2013best}, and the problem becomes more difficult in the presence of heterogeneous sensing, truncated sensing ranges, obstacle-induced infeasibility, and LoS occlusion.

From an optimization perspective, metaheuristic methods have been widely adopted for exposure-based path planning and related WSN optimization problems because they can handle highly nonlinear, discontinuous, and constrained search spaces without requiring gradient information \cite{houssein2024metaheuristic}. Among them, evolutionary algorithms (EAs) are especially attractive because they naturally operate on a population of candidate solutions and can be extended to \emph{multiobjective} optimization, making them well suited for approximating diverse Pareto-optimal trade-offs \cite{luong2025mobile,duc2025hybrid}. However, despite substantial progress in exposure-based path planning, most existing MaEP studies still formulate the problem primarily as a \emph{single-objective} exposure-maximization task. As a result, they only partially address the combined effects of heterogeneous sensing, obstacle-aware feasibility, and LoS occlusion, and they do not explicitly model the practical trade-off between sensing effectiveness and traversal efficiency.
This limitation is important in realistic surveillance scenarios. A path with extremely high accumulated exposure may also be excessively long, geometrically impractical, or undesirable for time-critical evaluation. Conversely, a short feasible path may fail to adequately reflect the strongest sensing regions of the network. Therefore, exposure maximization and path-length minimization are inherently conflicting objectives, and addressing them jointly is essential for producing practically meaningful solutions.

Motivated by these observations, this paper investigates an \emph{obstacle-aware multiobjective maximal exposure path planning} problem in IoT-enabled heterogeneous WSNs under a truncated omni-directional sensing model with LoS occlusion. We aim to determine a feasible side-to-side traversal path that avoids obstacles while simultaneously maximizing accumulated sensing exposure and minimizing path length. {To solve the problem, we develop \emph{OG-MOEA/D}, a problem-specific instantiation of MOEA/D in which a fixed-grid path representation and obstacle-aware feasibility handling are coupled with standard decomposition-based multiobjective search. The main contributions of this work are summarized as follows:}
\begin{itemize}
    \item {We formulate obstacle-aware maximal-exposure path planning as an exposure--length bi-objective problem under heterogeneous truncated sensing, LoS occlusion, side-to-side traversal, and polygonal obstacle constraints. The contribution is the joint formulation rather than any of these ingredients in isolation.}
    \item {We develop OG-MOEA/D by combining a fixed-dimensional path encoding and a reproducible adaptive feasibility-repair procedure with standard MOEA/D search. The standard MOEA/D decomposition, neighborhood collaboration, SBX, and polynomial mutation operators are not claimed as new.}
    \item {We strengthen the empirical validation with direct obstacle-constrained comparisons against MOEA/D and NSGA-II over 20 independent runs, explicit HV/IGD evaluation protocols, parameter-sensitivity analysis, and a grid-resolution study that quantifies the representational trade-off of the fixed-grid encoding.}
\end{itemize}

The remainder of this paper is organized as follows. Section \ref{sec:related-works} reviews related studies. Section \ref{sec:preliminaries-problem} introduces the sensing and obstacle models, defines the exposure measure, and formulates the OE-MEP problem as a bi-objective optimization task. Section \ref{sec:proposed-methodology} presents the proposed methodology, including solution representation, objective evaluation, and the optimization procedure. Section \ref{sec:results} reports the experimental setup and discusses the numerical results. Finally, Section \ref{sec:conclusion} concludes the paper and outlines future research directions.

\section{Related Works}\label{sec:related-works}

Barrier coverage has been a long-standing research topic in wireless sensor networks (WSNs) and IoT-enabled sensing systems because of its direct relevance to intrusion detection and surveillance in constrained environments. Existing studies can be broadly grouped into two complementary directions: \emph{(i) barrier construction} and \emph{(ii) penetration path planning} \cite{tripathi2018coverage,maheshwari2019survey}. The former aims to guarantee detectability for all side-to-side crossings, whereas the latter focuses on explicitly identifying representative traversal paths and evaluating their sensing characteristics. Since the maximal exposure path problem belongs to the second direction, this section briefly reviews both lines of research and then highlights the gap addressed in this work.

In the barrier-construction literature, the primary objective is to establish one or multiple sensing barriers such that any crossing path from one side of the monitored region to the opposite side is detected. This problem has been studied under a variety of sensing models, including directional, probabilistic, and heterogeneous sensing, often with additional requirements on energy efficiency or stronger forms of barrier coverage \cite{fan2019cost,ma2020exact,binh2024efficient}. For example, Fan et al.~\cite{fan2019cost} investigated cost-effective barrier construction in directional sensor networks by exploiting adjustable sensing ranges and cooperative probabilistic detection. Ma et al.~\cite{ma2020exact} proposed exact algorithms for barrier verification and construction with line-based deployed rotatable directional sensors. For heterogeneous wireless multimedia sensor networks, Binh et al.~\cite{binh2024efficient} developed a polynomial-time exact method for achieving $k$-strong barrier coverage. Although these studies provide important guarantees for surveillance quality, they focus on ensuring coverage for all feasible crossings rather than analyzing the properties of specific penetration paths.

In contrast, penetration path planning seeks an explicit side-to-side traversal path that satisfies or optimizes a given sensing criterion. Two canonical formulations are commonly considered: the \emph{minimal exposure path} (MiEP), which represents the most stealthy traversal, and the \emph{maximal exposure path} (MaEP), which characterizes the best-case sensing or detection capability of the network \cite{hong2017finding,wang2020generalized,van2023efficient}. MiEP has been widely studied as a tool for assessing surveillance vulnerabilities. Lee et al.~\cite{lee2013best} analyzed best- and worst-case coverage path problems and showed that exposure-path planning is computationally challenging, motivating the use of approximate and heuristic search methods. To better reflect realistic deployments, obstacle-aware MiEP variants have also been considered. In particular, Binh et al.~\cite{binh2022family} proposed an evolutionary framework for the minimal exposure path problem in irregular environments with obstacles, showing that evolutionary search can effectively handle obstacle-constrained continuous spaces.

{Liu et al.~\cite{liu2017obstacle} studied obstacle-avoidance minimal-exposure paths in heterogeneous WSNs, where the objective is to identify weakly monitored routes for vulnerability-oriented analysis. This is fundamentally different from the present maximal-exposure setting: our goal is to characterize best-case sensing strength while simultaneously limiting traversal length, producing a Pareto set rather than a single stealth-oriented route. The novelty of the present work therefore lies in the joint obstacle-aware maximal-exposure/length formulation under heterogeneous truncated sensing and LoS occlusion, rather than in obstacle handling or heterogeneous sensing considered in isolation.}
Compared with MiEP, MaEP has received relatively less attention, despite its importance for evaluating sensing effectiveness and stress-testing surveillance systems. Hong et al.~\cite{hong2017finding} studied best- and worst-case crossing paths in camera sensor networks and developed geometric algorithms that can achieve optimality under suitable convex partitions. Van et al.~\cite{van2023efficient} formulated MaEP using a sensing-intensity path integral and proposed a particle-swarm-inspired solution approach in homogeneous wireless ad-hoc sensor networks. More recently, Binh and Luong \cite{binh2025solving} investigated the maximal exposure path problem in heterogeneous directional WSNs and introduced an improved genetic algorithm to cope with heterogeneous sensing characteristics and the combinatorial nature of path search. These studies confirm the practical importance of MaEP and the suitability of metaheuristic search for such problems.

{From the multiobjective-optimization perspective, NSGA-II~\cite{deb2002nsga2}, MOEA/D~\cite{zhang2007moead}, and SPEA2~\cite{zitzler2001spea2} represent complementary dominance-, decomposition-, and strength-Pareto-based search paradigms that are widely used to approximate trade-off fronts. Related wireless-network studies have also optimized mobility jointly with communication or computing decisions: Sun et al.~\cite{sun2023uavfair} studied three-dimensional UAV trajectory design together with transmission scheduling for fair data collection, while Cheng et al.~\cite{cheng2025uavmec} considered joint 3D flight optimization and resource allocation for UAV-assisted mobile edge computing. These works demonstrate the broader importance of trajectory-aware multiobjective optimization in wireless systems. Their decision variables and performance objectives differ from OE-MEP, which focuses on obstacle-constrained sensing exposure and traversal length, but they provide relevant algorithmic and application context.}
Nevertheless, existing MaEP studies still exhibit several important limitations. First, most of them formulate the problem primarily as a \emph{single-objective} optimization task centered on exposure maximization alone. Second, the combined effects of heterogeneous sensing, obstacle-induced infeasibility, truncated sensing ranges, and line-of-sight (LoS) occlusion are only partially addressed in the literature. Third, practical traversal cost, especially path length, is often overlooked, even though it directly affects the usefulness and interpretability of the resulting penetration paths in realistic surveillance scenarios. As a consequence, prior formulations do not fully capture the trade-off between sensing effectiveness and traversal efficiency.

To address this gap, the present work formulates obstacle-aware maximal exposure path planning as a \emph{bi-objective} optimization problem in IoT-enabled heterogeneous sensor networks. The goal is to identify a side-to-side, obstacle-avoiding path that simultaneously \emph{maximizes accumulated sensing exposure} and \emph{minimizes path length} under truncated omni-directional sensing and LoS occlusion. This formulation leads naturally to a multiobjective evolutionary optimization perspective, where the aim is not to return a single path, but to approximate a Pareto set of exposure--length trade-offs under realistic sensing and geometric constraints.

\section{Preliminaries and Problem Formulation}\label{sec:preliminaries-problem}

\subsection{Preliminaries on sensing models and obstacles}\label{AA}

Sensor nodes in WSNs may differ substantially in both their theoretical sensing assumptions and their physical capabilities. As a result, sensing models with different levels of fidelity are often adopted depending on the target application and deployment conditions. A common characteristic shared by many sensing models is that sensing quality---such as detection strength, confidence, or signal intensity---decreases as the distance from the sensor increases. Therefore, realistic path-planning formulations should account not only for the spatial locations of sensors, but also for how sensing strength varies across the monitored region.

In this work, we adopt the \emph{truncated omni-directional} sensing model, which explicitly captures the finite sensing range of practical devices. Unlike idealized attenuated models whose influence extends indefinitely, the truncated model assumes that each sensor has a hard maximum sensing radius $r$, beyond which its contribution becomes zero. Specifically, the sensing intensity contributed by a sensor $s$ at position $P$ to a target point $T$ is defined as
\begin{equation}
{f(s,T)}=
\begin{cases}
\dfrac{C}{1 + \big(d(P,T)\big)^{\lambda}}, & \text{if } d(P,T) \le r,\\[6pt]
0, & \text{otherwise},
\end{cases}
\label{eq:f2}
\end{equation}
where $d(P,T)$ denotes the Euclidean distance between the sensor location $P$ and the target point $T$, while $C$ and $\lambda$ are attenuation parameters determined by the sensor characteristics. Hence, the model preserves the distance-dependent decay of sensing quality within the sensing radius while enforcing a realistic finite coverage boundary.

\vspace{6pt}
\paragraph*{Accumulative sensing intensity.}
Since an intruder may simultaneously be influenced by multiple sensors, the sensing intensity at a point should be evaluated collectively rather than individually. Let $\mathcal{S}=\{s_1,s_2,\ldots,s_N\}$ be the set of deployed sensors. The \emph{accumulative sensing intensity} at a point $T$, denoted by $\mathsf{ASI}(\mathcal{S},T)$, is defined as
\begin{equation}
\mathsf{ASI}(\mathcal{S},T)=\sum_{i=1}^{N} f(s_i,T),
\label{eq:f3}
\end{equation}
where $f(s_i,T)$ is computed using the truncated omni-directional model in \eqref{eq:f2}. This aggregate quantity reflects the total sensing influence experienced at location $T$ and serves as the basis for defining path exposure later in this section.

\subsection{Obstacle model and occlusion}

In realistic environments, the ROI often contains physical obstacles such as buildings, lakes, dense vegetation, or restricted zones. These obstacles affect the problem in two distinct ways. First, they constrain feasibility by limiting valid sensor placement and admissible traversal paths. Second, they alter sensing itself by blocking line-of-sight and thereby reducing or nullifying sensing contributions in occluded regions. Accurately modeling both effects is essential for obstacle-aware exposure path planning.

Let $\mathcal{O}$ denote the set of obstacles in the ROI. Each obstacle $O \in \mathcal{O}$ is modeled as a convex polygon in the two-dimensional plane and is represented by an ordered set of vertices
\begin{equation}
G_O=\left\{P_1^O,P_2^O,\ldots,P_M^O\right\},
\label{eq:obstacle_vertices}
\end{equation}
where consecutive vertices define the polygon boundary, and the final vertex is connected back to the first one. The interior of the obstacle, denoted by $\mathrm{int}(O)$, is regarded as forbidden space. Therefore, sensors cannot be deployed inside an obstacle, and any feasible traversal path must avoid all obstacle interiors. These feasibility requirements are written as
\begin{equation}
P_i \notin \mathrm{int}(O), \quad \forall s_i \in \mathcal{S},\ \forall O \in \mathcal{O},
\label{eq:sensor_forbidden}
\end{equation}
and, for any candidate intrusion path $\wp:[0,\tau]\rightarrow\mathbb{R}^2$,
\begin{equation}
\wp(t)\notin \mathrm{int}(O), \quad \forall t\in[0,\tau],\ \forall O \in \mathcal{O},
\label{eq:path_forbidden}
\end{equation}
where $\tau$ denotes the traversal time.

Besides constraining geometry, obstacles are also assumed to fully absorb sensing signals. In other words, a point can only be sensed by a sensor if a line-of-sight (LoS) connection exists between them. Formally, for a sensor $s_i$ located at $P_i$ and a point $P$ in the ROI, if the line segment $\overline{P_iP}$ intersects the interior of any obstacle, then the sensing contribution of $s_i$ at $P$ is nullified:
\begin{equation}
\text{if } \exists O \in \mathcal{O} : \overline{P_iP}\cap \mathrm{int}(O)\neq \varnothing,
\text{ then } f(s_i,P)=0.
\label{eq:occlusion_rule}
\end{equation}

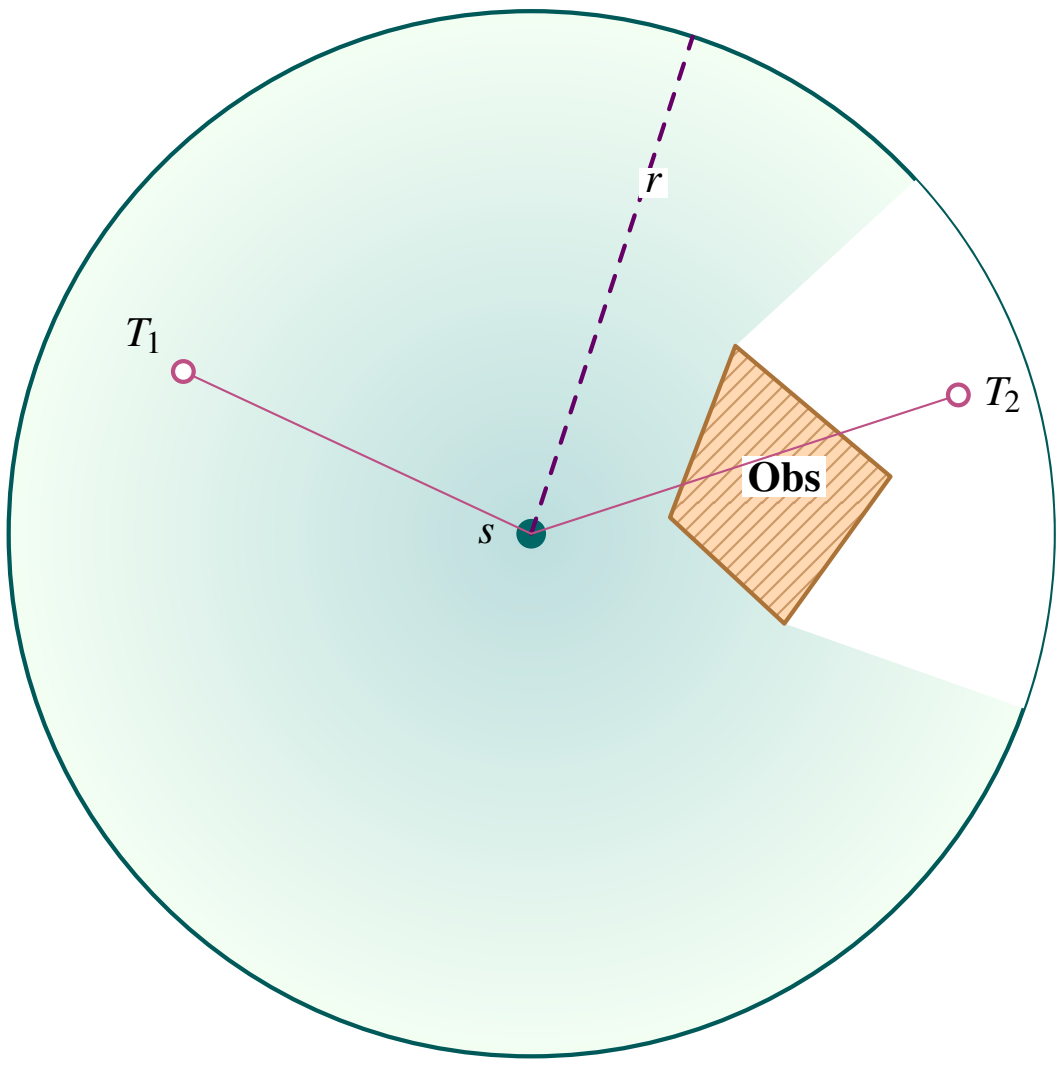
\begin{figure}[htbp]
    \centering
    \begin{tikzpicture}[>=Stealth, scale=1.08, line cap=round, line join=round]

        \coordinate (S) at (0,0);
        \def\R{3.2}

        \coordinate (O1) at (1.25,1.15);
        \coordinate (O2) at (2.20,0.35);
        \coordinate (O3) at (1.55,-0.55);
        \coordinate (O4) at (0.85,0.10);

        \shade[inner color=teal!30, outer color=green!5, opacity=0.85] (S) circle (\R);
        \draw[teal!70!black, thick] (S) circle (\R);

        \path let \p1 = ($(O1)-(S)$), \n1 = {atan2(\y1,\x1)} in
            coordinate (A) at (\n1:\R);
        \path let \p1 = ($(O3)-(S)$), \n1 = {atan2(\y1,\x1)} in
            coordinate (B) at (\n1:\R);

        \fill[white]
            (O1) -- (A)
            arc[start angle={atan2(1.15,1.25)}, end angle={atan2(-0.55,1.55)}, radius=\R]
            -- (O3) -- (O2) -- cycle;

        \fill[orange!30] (O1) -- (O2) -- (O3) -- (O4) -- cycle;
        \fill[pattern=north east lines, pattern color=brown!80]
            (O1) -- (O2) -- (O3) -- (O4) -- cycle;
        \draw[brown!90!black, thick] (O1) -- (O2) -- (O3) -- (O4) -- cycle;
        \node[font=\footnotesize\bfseries, fill=white, inner sep=1pt] at (1.55,0.35) {Obs};

        \filldraw[teal!80!black] (S) circle (2.4pt);
        \node[left=4pt] at (S) {\textbf{$s$}};

        \draw[dashed, thick, violet!80!black] (S) -- (72:\R);
        \node[fill=white, inner sep=1pt] at (0.75,2.15) {$r$};

        \coordinate (T1) at (155:2.35);
        \draw[magenta!75!black, thin] (S) -- (T1);
        \filldraw[white, draw=magenta!75!black, thick] (T1) circle (1.8pt);
        \node[above left=1pt] at (T1) {$T_1$};

        \coordinate (T2) at (18:2.75);
        \draw[magenta!75!black, thin] (S) -- (T2);
        \filldraw[white, draw=magenta!75!black, thick] (T2) circle (1.8pt);
        \node[right=2pt] at (T2) {$T_2$};

    \end{tikzpicture}
    \caption{Illustration of truncated sensing with obstacle-induced occlusion. The blue disk represents the nominal sensing region of sensor $s$. Point $T_1$ is within range and visible to the sensor, while point $T_2$ is within the nominal radius but lies behind the obstacle and therefore receives zero sensing contribution according to \eqref{eq:occlusion_rule}.}
    \label{fig:sensor_obstacle_model}
\end{figure}

Fig.~\ref{fig:sensor_obstacle_model} illustrates this sensing behavior. The blue circular region represents the nominal sensing range of the sensor, while the hatched polygon denotes an obstacle. Although a point may lie within the sensing radius, it is still undetectable if it lies in the shadow region behind the obstacle. Thus, obstacle-aware exposure depends jointly on distance attenuation, finite sensing range, and LoS visibility.

\subsection{Path exposure}

The exposure of a path quantifies how strongly an intruder is sensed while moving through the monitored field. Intuitively, a path has high exposure if it passes through regions with high accumulative sensing intensity for a sufficiently long distance. Let $\wp$ denote a continuous traversal path. The exposure induced by the sensor set $\mathcal{S}$ is defined as the line integral of the accumulative sensing intensity along the path:
\begin{equation}
\mathsf{Exp}(\mathcal{S},\wp)=\int_{T\in\wp}\mathsf{ASI}(\mathcal{S},T)\,ds,
\label{eq:f4}
\end{equation}
where $ds$ denotes an infinitesimal arc-length element on $\wp$.

Because the exposure function may be nonlinear and non-differentiable in the presence of truncated sensing and obstacle-induced occlusion, direct analytical evaluation is generally intractable. We therefore approximate \eqref{eq:f4} numerically by discretizing the path into $K$ sample points $\{T_j\}_{j=1}^{K}$ with arc-length spacing $\Delta s$. When $\Delta s$ is sufficiently small, the exposure can be approximated by
\begin{equation}
\mathsf{Exp}(\mathcal{S},\wp)\approx \sum_{j=1}^{K}\mathsf{ASI}(\mathcal{S},T_j)\,\Delta s.
\label{eq:f5}
\end{equation}
Substituting \eqref{eq:f3} into \eqref{eq:f5} yields
\begin{equation}
\mathsf{Exp}(\mathcal{S},\wp)\approx
\sum_{j=1}^{K}\mathsf{ASI}(\mathcal{S},T_j)\,\Delta s
=
\sum_{j=1}^{K}\sum_{i=1}^{N}f(s_i,T_j)\,\Delta s,
\label{eq:f6}
\end{equation}
where $N$ is the number of sensors in $\mathcal{S}$. In the experiments, the sensing function $f(s_i,T_j)$ is computed using the truncated omni-directional model in \eqref{eq:f2} together with the LoS occlusion rule in \eqref{eq:occlusion_rule}.

\subsection{Problem formulation}

We now formalize the obstacle-aware maximal exposure path planning problem, abbreviated as OE-MEP. Consider a rectangular ROI $\Omega$ with width $W$ and height $H$, containing a set of deployed truncated omni-directional sensors
$\mathcal{S}=\{s_1,s_2,\ldots,s_N\}$ and a set of polygonal obstacles
$\mathcal{O}=\{O_1,O_2,\ldots,O_J\}$. Unlike classical maximal-exposure-path settings with fixed start and end points, we consider \emph{side-to-side traversal}: an intruder enters the ROI at a point $B$ on a designated boundary and exits at a point $E$ on the opposite boundary.

The objective is to determine a continuous and collision-free penetration path $\wp$ from $B$ to $E$ such that the path avoids all obstacles, satisfies a maximum allowable length, and achieves a desirable trade-off between two conflicting goals: maximizing the accumulated sensing exposure and minimizing the geometric path length. {The length objective prevents exposure from being increased simply through unnecessarily long or highly oscillatory trajectories and yields an interpretable surveillance-strength/traversal-efficiency trade-off; it does not convert maximal exposure into an evasive-intrusion model.}
\newpage
\noindent\textbf{Input:}
\begin{itemize}
    \item $W,H$: the width and height of the ROI $\Omega$;
    \item $L_{\max}$: the maximum allowable length of the penetration path $\wp$;
    \item $N$: the number of sensors;
    \item for each sensor $s_i$ $(i=1,2,\ldots,N)$:
    \begin{itemize}
        \item $(x_i,y_i)$: the location of $s_i$;
        \item $r_i$: the sensing radius of $s_i$;
        \item $C_i,\lambda_i$: the attenuation parameters of $s_i$.
    \end{itemize}
    \item $J$: the number of obstacles;
    \item for each obstacle $O_l$ $(l=1,2,\ldots,J)$:
    \begin{itemize}
        \item $G_{O_l}$: the ordered list of vertices defining $O_l$.
    \end{itemize}
\end{itemize}

\noindent\textbf{Output:}  
An ordered set of points $L_{\wp}$ in $\Omega$ defining a continuous path $\wp$ that connects an entry point $B$ on the starting edge to an exit point $E$ on the opposite edge.

\noindent\textbf{Objectives:}  
OE-MEP is formulated as a bi-objective optimization problem with two conflicting criteria:
\begin{itemize}
    \item \textbf{Objective 1:} maximize the total exposure accumulated along the path;
    \item \textbf{Objective 2:} minimize the total geometric path length.
\end{itemize}
For a minimization-based formulation, the first objective is written as the negative exposure:
\begin{align}
f_1(\wp) &= -\mathsf{Exp}(\mathcal{S},\wp)
\approx -\sum_{j=1}^{K}\sum_{i=1}^{N}f(s_i,T_j)\,\Delta s,
\label{eq:obj1}\\
f_2(\wp) &= L(\wp)=\sum_{j=1}^{K-1}\left\|\overrightarrow{T_jT_{j+1}}\right\|.
\label{eq:obj2}
\end{align}

\noindent\textbf{Constraints:}  
A feasible path must remain inside the ROI, connect the designated entry and exit boundaries, avoid all obstacle interiors, and satisfy the maximum-length budget.

Accordingly, the OE-MEP problem is formulated as
\begin{mini!}|l|<b>
{\wp \subseteq \Omega} {\mathbf{f}(\wp)=\big\{f_1(\wp),f_2(\wp)\big\}}
{\label{OEMEP}}{}
\addConstraint{B \in \partial\Omega_{\text{start}},\quad E \in \partial\Omega_{\text{end}} \label{eq:con_endpoints}}
\addConstraint{\wp \cap \mathrm{int}(O_l)=\varnothing,\quad \forall O_l \in \mathcal{O} \label{eq:con_obstacles}}
\addConstraint{L(\wp)\le L_{\max} \label{eq:con_length}}
\end{mini!}

This formulation couples sensing quality and traversal efficiency under obstacle and LoS constraints. As a result, the goal is not merely to find a single highly exposed path, but to characterize the trade-off surface between exposure and path length for obstacle-aware side-to-side crossings in heterogeneous WSNs.
{Equation~\eqref{OEMEP} defines the general continuous OE-MEP problem over the feasible path set $\mathcal{P}$. The fixed-grid implementation considered in this paper searches the restricted family $\mathcal{P}^{\uparrow}_M\subset\mathcal{P}$ of $x$-monotone polylines induced by $M$ equal horizontal intervals. Consequently, the reported solutions approximate the Pareto set within $\mathcal{P}^{\uparrow}_M$; we do not claim global optimality over unrestricted continuous paths that may require horizontal backtracking.}

\section{Proposed Algorithm}\label{sec:proposed-methodology}

Evolutionary algorithms (EAs) are well suited to complex optimization problems with nonlinear, nonconvex, and highly constrained search spaces because they do not require gradient information and can explore diverse candidate solutions in parallel. These properties are attractive for OE-MEP, where obstacle avoidance, LoS occlusion, and path-length constraints create a rugged search landscape.
\begin{algorithm}[t]
\caption{OG-MOEA/D for Obstacle-Aware Maximal Exposure Path Planning}
\label{alg:og_moead}
\begin{algorithmic}[1]
\STATE \textbf{Input:} ROI size $(W,H)$, sensor set $\mathcal{S}$, obstacle set $\mathcal{O}$, maximum path length $L_{\max}$, number of grid intervals $M$, population size $N_p$, neighborhood size $T$, maximum generations $G$, crossover rate $p_c$, mutation rate $p_m$, distribution indices $\eta_c,\eta_m$, initialization limit $K_{\rm init}=50$, final repair limit $K_{\rm rep}=100$
\STATE \textbf{Output:} Approximate Pareto set $\mathcal{EP}$

\STATE Generate uniformly distributed weight vectors $\Lambda=\{\boldsymbol{\lambda}^1,\dots,\boldsymbol{\lambda}^{N_p}\}$
\STATE Construct neighborhood sets $B(k)$ for each subproblem $k$
\STATE $\mathcal{P} \leftarrow \emptyset$

\FOR{$k=1$ to $N_p$}
    \STATE $\mathrm{success}\leftarrow\mathrm{false}$
    \FOR{$q=1$ to $K_{\rm init}$}
        \STATE Randomly generate $Y^k=\{y_0,\dots,y_M\}$ with $y_i\sim\mathcal{U}[0,H]$
        \IF{$\wp(Y^k)$ is feasible and $L(\wp(Y^k))\le L_{\max}$}
            \STATE $\mathrm{success}\leftarrow\mathrm{true}$; \quad \textbf{break}
        \ENDIF
    \ENDFOR
    \IF{$\mathrm{success}=\mathrm{false}$}
        \STATE Declare the current initialization attempt unsuccessful and restart the run; no unbounded rejection loop is used.
    \ENDIF
    \STATE $\mathcal{P}\leftarrow\mathcal{P}\cup\{Y^k\}$; \quad Evaluate $f_1(\wp(Y^k))$ and $f_2(\wp(Y^k))$
\ENDFOR

\STATE Initialize ideal point $z^*=(z_1^*,z_2^*)$ by \eqref{eq:ideal_point}, initialize external archive $\mathcal{EP}$

\FOR{$g=1$ to $G$}
    \FOR{$k=1$ to $N_p$}
        \STATE Randomly select two parent indices $r_1,r_2 \in B(k)$
        \STATE Let $Y^{p_1} \leftarrow \mathcal{P}[r_1]$ and $Y^{p_2} \leftarrow \mathcal{P}[r_2]$
        \STATE {Initialize the offspring as $\widetilde{Y}\leftarrow Y^{p_1}$}

        \IF{rand() $< p_c$}
            \STATE Generate offspring $\widetilde{Y}$ from $Y^{p_1}$ and $Y^{p_2}$ by SBX using $\eta_c$
        \ENDIF

        \STATE Apply polynomial mutation to $\widetilde{Y}$ with probability $p_m$ and index $\eta_m$
        
        \STATE Repair $\widetilde{Y}$ using the adaptive repair mechanism:
    
        \STATE Perform final validation for all line segments in $\wp(\widetilde{Y})$
\IF{$\wp(\widetilde{Y})$ is feasible and $L(\wp(\widetilde{Y})) \le L_{\max}$}
            \STATE Evaluate $f_1(\wp(\widetilde{Y}))$ and $f_2(\wp(\widetilde{Y}))$
            \FOR{$m=1$ to $2$}
                \STATE $z_m^* \leftarrow \min\{z_m^*, f_m(\wp(\widetilde{Y}))\}$
            \ENDFOR

            \FOR{each $j \in B(k)$}
                \STATE Compute
                $g^{te}(\widetilde{Y}\mid \boldsymbol{\lambda}^j,z^*)$, 
                $g^{te}(Y^j\mid \boldsymbol{\lambda}^j,z^*)
               $.
                \IF{$g^{te}(\widetilde{Y}\mid \boldsymbol{\lambda}^j,z^*) < g^{te}(Y^j\mid \boldsymbol{\lambda}^j,z^*)$}
                    \STATE Replace $Y^j$ in $\mathcal{P}$ by $\widetilde{Y}$
                \ENDIF
            \ENDFOR

            \STATE Update archive $\mathcal{EP}$ with $\widetilde{Y}$ and remove dominated solutions
        \ENDIF
    \ENDFOR
\ENDFOR

\STATE \textbf{return} $\mathcal{EP}$
\end{algorithmic}
\end{algorithm}

OG-MOEA/D is a problem-specific instantiation of MOEA/D rather than a new decomposition principle. The decomposition strategy, neighborhood mating/update, simulated binary crossover (SBX), and polynomial mutation follow standard MOEA/D practice~\cite{zhang2007moead}. The problem-specific additions are (i) the fixed-grid representation of side-to-side paths, (ii) exposure/length evaluation under truncated sensing and LoS occlusion, and (iii) obstacle- and length-aware feasibility checking and repair. These problem-specific modules are conceptually reusable with another population-based multiobjective optimizer by replacing the MOEA/D-specific scalarization, mating, and environmental-update steps. In this work, MOEA/D is used as the search engine because its decomposition structure maintains search pressure along different exposure--length trade-off directions.
The pseudo-code of OG-MOEA/D is presented in Algorithm~\ref{alg:og_moead}. The remainder of this section describes the solution representation, initialization, decomposition strategy, variation operators, feasibility repair, and neighborhood-based update mechanism.

\subsection{Solution representation and initialization}

A candidate solution encodes a side-to-side penetration path within the ROI $\Omega=[0,W]\times[0,H]$. Rather than representing a path by a variable-length sequence of free $(x,y)$ waypoints, which would complicate crossover, mutation, and feasibility control, OG-MOEA/D adopts a fixed-grid representation along the horizontal axis. This design yields a compact fixed-length genotype, simplifies operator design, and makes it easier to compare and update solutions across subproblems.

Specifically, the horizontal axis is discretized into $M$ uniform intervals with step size $\Delta x = W/M$, giving the grid
\begin{equation}
X=\{x_0,x_1,\ldots,x_M\}, \qquad x_i=i\,\Delta x .
\label{eq:x_grid}
\end{equation}
A solution stores only the corresponding vertical coordinates
\begin{equation}
Y=\{y_0,y_1,\ldots,y_M\}, \qquad y_i\in[0,H],
\label{eq:y_genotype}
\end{equation}
which define the polyline path
\begin{equation}
\wp(Y)=\bigcup_{i=0}^{M-1}\overline{P_iP_{i+1}}, \qquad P_i=(x_i,y_i).
\label{eq:path_polyline}
\end{equation}

This representation has two practical advantages. First, the search space remains continuous in the vertical dimension, allowing flexible path shaping. Second, the path complexity is controlled by the grid resolution $M$, which provides a convenient balance between representation fidelity and computational cost.

{Because $x_i=i\Delta x$ is strictly increasing, every representable path is $x$-monotone. Let $\mathcal{P}^{\uparrow}_M$ denote this restricted family. The representation may therefore exclude feasible or optimal unrestricted paths that require horizontal backtracking around complex obstacles. We retain it because it yields a fixed-dimensional real-valued genotype and stable variation/repair operations, and we quantify the resolution trade-off experimentally in Section 5.}

A solution $Y$ is feasible only if all gene values remain inside the ROI and the induced polyline avoids obstacle interiors:
\begin{equation}
0 \le y_i \le H, \qquad \forall i=0,\ldots,M,
\label{eq:y_bounds}
\end{equation}
\begin{equation}
\wp(Y)\cap \mathrm{int}(O_\ell)=\varnothing, \qquad \forall O_\ell\in\mathcal{O}.
\label{eq:obs_constraint_path}
\end{equation}

Let $N_p$ denote the population size. The initial population
$\mathcal{P}=\{Y^1,Y^2,\ldots,Y^{N_p}\}$ is generated by rejection-based random sampling. For each candidate, a vertical coordinate $y_i$ is sampled uniformly from $[0,H]$ at every fixed grid point $x_i$, producing a genotype vector $Y$. The resulting path is then checked against the feasibility constraints in \eqref{eq:y_bounds}, \eqref{eq:obs_constraint_path}, and the maximum-length constraint defined in Section~\ref{sec:preliminaries-problem}. {To avoid an unbounded rejection loop, at most $K_{\rm init}=50$ trials are used for each individual. If no feasible sample is obtained within this limit, the current initialization attempt is declared unsuccessful and the run is restarted. Thus, initialization always has an explicit finite safeguard.}

After initialization, the ideal reference point
\begin{equation}
z^*=(z_1^*,z_2^*)
\label{eq:ideal_point}
\end{equation}
is constructed from the best objective values observed in the initial population, where
\begin{equation}
z_m^*=\min_{Y\in\mathcal{P}} f_m(\wp(Y)), \qquad m\in\{1,2\}.
\label{eq:ideal_point_update}
\end{equation}

\subsection{Problem decomposition and neighborhood collaboration}

Instead of optimizing the bi-objective vector $\mathbf{f}(\wp)=\{f_1(\wp),f_2(\wp)\}$ directly, OG-MOEA/D decomposes the original problem into $N_p$ scalar subproblems that are optimized simultaneously. Let
\begin{equation}
\Lambda=\{\boldsymbol{\lambda}^1,\boldsymbol{\lambda}^2,\ldots,\boldsymbol{\lambda}^{N_p}\},
\qquad
\boldsymbol{\lambda}^k=(\lambda_1^k,\lambda_2^k),
\label{eq:weight_vectors}
\end{equation}
be a set of uniformly distributed weight vectors. Each weight vector defines one search direction in the objective space.

In this work, we employ the Tchebycheff decomposition strategy. For the $k$-th subproblem, the scalar objective value of a solution $Y$ is defined as
\begin{equation}
g^{te}(Y \mid \boldsymbol{\lambda}^k,z^*)
=
\max_{m\in\{1,2\}}
\left\{
\lambda_m^k\,\big|f_m(\wp(Y))-z_m^*\big|
\right\}.
\label{eq:tchebycheff}
\end{equation}
Minimizing \eqref{eq:tchebycheff} encourages the solution to approach the ideal point along a direction controlled by $\boldsymbol{\lambda}^k$, thereby promoting a diverse approximation of the Pareto front.

A key strength of MOEA/D is that neighboring subproblems are allowed to share information. The underlying intuition is that subproblems with similar weight vectors tend to have similar optimal solutions. Therefore, instead of evolving each subproblem independently, OG-MOEA/D exploits local collaboration through neighborhood-based mating and replacement.

For each subproblem $k$, we define a neighborhood set
\begin{equation}
B(k)=\{j_1,j_2,\ldots,j_T\}\subseteq\{1,2,\ldots,N_p\},
\label{eq:neighborhood}
\end{equation}
where $T$ is the neighborhood size and the selected indices correspond to the $T$ closest weight vectors to $\boldsymbol{\lambda}^k$ in Euclidean distance. This neighborhood serves two purposes:
\begin{itemize}
    \item \textbf{Mating selection:} parents used to generate an offspring for subproblem $k$ are preferentially selected from $B(k)$.
    \item \textbf{Population update:} an improved offspring may replace the incumbent solutions of multiple neighboring subproblems if it yields better scalar values.
\end{itemize}
This mechanism improves search efficiency by encouraging local information reuse while still maintaining multiple search directions across the population.

\subsection{Variation operators}

Because each solution is represented by a real-valued vector of vertical coordinates, OG-MOEA/D employs real-coded variation operators that are well matched to continuous path refinement.

\paragraph*{Simulated Binary Crossover (SBX).}
Given two parent solutions $Y^{p_1}$ and $Y^{p_2}$ selected from the neighborhood, SBX generates offspring by blending the corresponding height values at each grid point. For gene $i$, the offspring values are computed as
\begin{equation}
y_i^{c_1}
=
\frac{1}{2}
\Big[
(1+\beta_i)y_i^{p_1}
+
(1-\beta_i)y_i^{p_2}
\Big],
\label{eq:sbx1}
\end{equation}
\begin{equation}
y_i^{c_2}
=
\frac{1}{2}
\Big[
(1-\beta_i)y_i^{p_1}
+
(1+\beta_i)y_i^{p_2}
\Big],
\label{eq:sbx2}
\end{equation}
where the spread factor $\beta_i$ is determined from a random variable $u\sim\mathcal{U}[0,1]$:
\begin{equation}
\beta_i=
\begin{cases}
(2u)^{\frac{1}{\eta_c+1}}, & \text{if } u\le 0.5, \\[6pt]
\left(\dfrac{1}{2(1-u)}\right)^{\frac{1}{\eta_c+1}}, & \text{otherwise}.
\end{cases}
\label{eq:sbx_beta}
\end{equation}
Here, $\eta_c$ is the crossover distribution index. Larger values of $\eta_c$ generate offspring closer to the parents and favor local exploitation, whereas smaller values encourage broader exploration. After crossover, each offspring gene is clipped to the interval $[0,H]$ to satisfy the bound constraint in \eqref{eq:y_bounds}.

\paragraph*{Polynomial mutation.}
To maintain diversity and reduce the risk of premature convergence, polynomial mutation is applied independently to each gene with probability $p_m$. For a mutated gene $y_i$, the perturbed value is given by
\begin{equation}
y'_i=
\begin{cases}
y_i+\delta_i\,(y_i-y_{\min}), & \text{if } \delta_i<0, \\[4pt]
y_i+\delta_i\,(y_{\max}-y_i), & \text{otherwise},
\end{cases}
\label{eq:pm1}
\end{equation}
where $y_{\min}=0$, $y_{\max}=H$, and the mutation perturbation $\delta_i$ is computed from $r\sim\mathcal{U}[0,1]$ as
\begin{equation}
\delta_i=
\begin{cases}
(2r)^{\frac{1}{\eta_m+1}}-1, & \text{if } r<0.5, \\[6pt]
1-\big[2(1-r)\big]^{\frac{1}{\eta_m+1}}, & \text{otherwise}.
\end{cases}
\label{eq:pm2}
\end{equation}
The parameter $\eta_m$ controls the mutation amplitude. This operator is particularly useful for locally adjusting path heights so that the search can progressively refine feasible corridors and explore nearby exposure-improving alternatives.

\subsection{Adaptive repair mechanism}

{
Offspring generated by crossover and mutation may violate geometric constraints, especially near obstacle boundaries or narrow feasible passages. Simply discarding such solutions immediately would waste useful inherited structure and reduce search efficiency. To address this issue, OG-MOEA/D incorporates an adaptive repair mechanism that attempts to restore feasibility while preserving as much of the original path shape as possible.

The repair process scans the genes of an offspring sequentially. {A waypoint is locally invalid when it lies outside the ROI or in an obstacle interior; segment-level intersections are checked during final polyline validation. We therefore avoid the ambiguous term ``unsafe local configuration.''} If a point $P_i=(x_i,y_i)$ is invalid, the algorithm repairs it through a hierarchical strategy:

\begin{itemize}
    \item \textbf{Local perturbation schedule:} The algorithm applies a random vertical displacement $\delta \sim \mathcal{U}[-r, r]$ to find a feasible coordinate. The search radius $r$ is progressively expanded following a predefined multi-scale schedule: $r \in \{2, 5, 10, 20, 40, 80, H\}$, where $H$ is the total height of the environment. For each intermediate radius level ($r \le 80$), the algorithm makes exactly 10 attempts to find a collision-free position before escalating to the next level.
    
    \item \textbf{Global search and failure handling:} {At the final radius level $r=H$, at most $K_{\rm rep}=100$ full-height perturbation trials are performed using the same sampling rule $\delta\sim\mathcal{U}[-H,H]$. If none of these trials yields a locally valid coordinate, the repair returns failure immediately. Otherwise, after all locally invalid waypoints have been processed, the complete polyline is revalidated; if any segment intersects an obstacle or the resulting path violates the domain or length constraint, the repair also returns failure and the offspring is discarded.}
    
    \item \textbf{Final validation and failure handling:} After repairing individual problematic genes, the full polyline must be re-evaluated. Although individual waypoints may now be obstacle-free, the line segments connecting consecutive points might still intersect obstacle interiors. If any segment violates this condition, the repair function returns a null state. In the main evolutionary loop, an offspring that triggers this failure state is entirely discarded. It is not evaluated and cannot replace any existing neighboring solutions, thereby strictly preventing the proliferation of invalid configurations.
\end{itemize}

{This explicit schedule makes the repair procedure reproducible under the stated implementation settings.} It effectively biases the search toward feasible regions without completely destroying the topological structure inherited from parent solutions. Small local corrections (at early radius levels) often suffice to move a path around an obstacle boundary, while the strict failure handling mechanism prevents the population from stagnating in severely constrained or inescapable local topologies.
}

\subsection{Neighborhood-based selection and update}

After variation and repair, the resulting offspring $\widetilde{Y}$ is evaluated using the bi-objective functions defined in Section~\ref{sec:preliminaries-problem}. The ideal point is then updated as
\begin{equation}
z_m^* \leftarrow \min\left\{z_m^*,\, f_m(\wp(\widetilde{Y}))\right\},
\qquad m\in\{1,2\}.
\label{eq:update_ideal}
\end{equation}

The offspring is subsequently used to update the neighborhood of the current subproblem. For each $j\in B(k)$, OG-MOEA/D compares the scalarized value of $\widetilde{Y}$ with that of the incumbent solution $Y^j$ under the weight vector $\boldsymbol{\lambda}^j$. If
\begin{equation}
g^{te}(\widetilde{Y}\mid \boldsymbol{\lambda}^j,z^*)
<
g^{te}(Y^j\mid \boldsymbol{\lambda}^j,z^*),
\label{eq:update_rule}
\end{equation}
then $Y^j$ is replaced by $\widetilde{Y}$.

This update rule allows a single high-quality offspring to improve several nearby subproblems at once, which accelerates the propagation of useful path structures across neighboring trade-off regions. As a result, OG-MOEA/D combines decomposition-based diversity preservation with effective local information sharing, making it well suited to obstacle-aware exposure path planning in heterogeneous WSNs.

\section{Numerical Results}\label{sec:results}

\subsection{Experimental settings}
All experiments were implemented in Python~3.13 and executed on a Windows~11 (25H2) workstation equipped with a 1.3~GHz 10-core CPU and 16~GB RAM. We considered four deployment topologies containing 50, 100, 150, and 200 sensors, respectively. In each topology, sensors were uniformly distributed over a rectangular ROI with width $W=400$ and height $H=200$. {Sensor heterogeneity is modeled at the individual-sensor level: for each sensor $i$, the sensing radius and attenuation parameters are sampled independently as
$r_i\sim\mathcal{U}[5,50]$, $C_i\sim\mathcal{U}[0.5,2]$, and $\lambda_i\sim\mathcal{U}[1.5,3]$. These ranges define the explicitly heterogeneous sensing regime evaluated throughout this study and avoid the homogeneous common-parameter setting used in the original description. Our experiments are intended to evaluate the optimization method under sensor-level heterogeneity; they do not attempt to estimate sensitivity to a separate ``heterogeneity-strength'' factor. We therefore restrict our claims to this stated heterogeneous regime and leave a systematic factorized study of heterogeneity magnitude to future work.} The maximum allowable path length was fixed to $L_{\max}=2000$.

{The default path grid uses $M=80$ horizontal intervals and therefore $n=M+1=81$ encoded vertical coordinates, giving $\Delta x=W/M=5$.} OG-MOEA/D was run for $G=100$ generations with population size $N_p=100$ and neighborhood size $T=15$. The crossover and mutation rates were set to $p_c=0.9$ and $p_m=0.4$, respectively, while the SBX and polynomial mutation operators used distribution indices $\eta_c=10$ and $\eta_m=10$. Each environment contained $J=5$ convex polygonal obstacles, and the obstacle coverage ratio was fixed to $oc=0.2$.

{For the main multiobjective comparison, each method was executed for 20 independent stochastic runs using consecutive integer seeds $0,\ldots,19$. All three methods used the same four topology definitions, population size $N_p=100$, generation limit $G=100$, obstacle-constrained evaluation protocol, and termination schedule. This corresponds to 10,000 offspring-generation opportunities per run, in addition to initialization. Wall-clock measurements are end-to-end implementation-level measurements on the same workstation. The seed values are reported for reproducibility; the run-level performance samples are compared using the unpaired non-parametric test. The parameter- and grid-resolution sensitivity analyses use five independent seeds per tested setting, as stated in the corresponding subsections.}

\subsection{Experimental results and discussions}

\subsubsection{Analysis of convergence and diversity using hypervolume}
\begin{figure}[htbp]
    \centering
\includegraphics[width=0.9\linewidth]{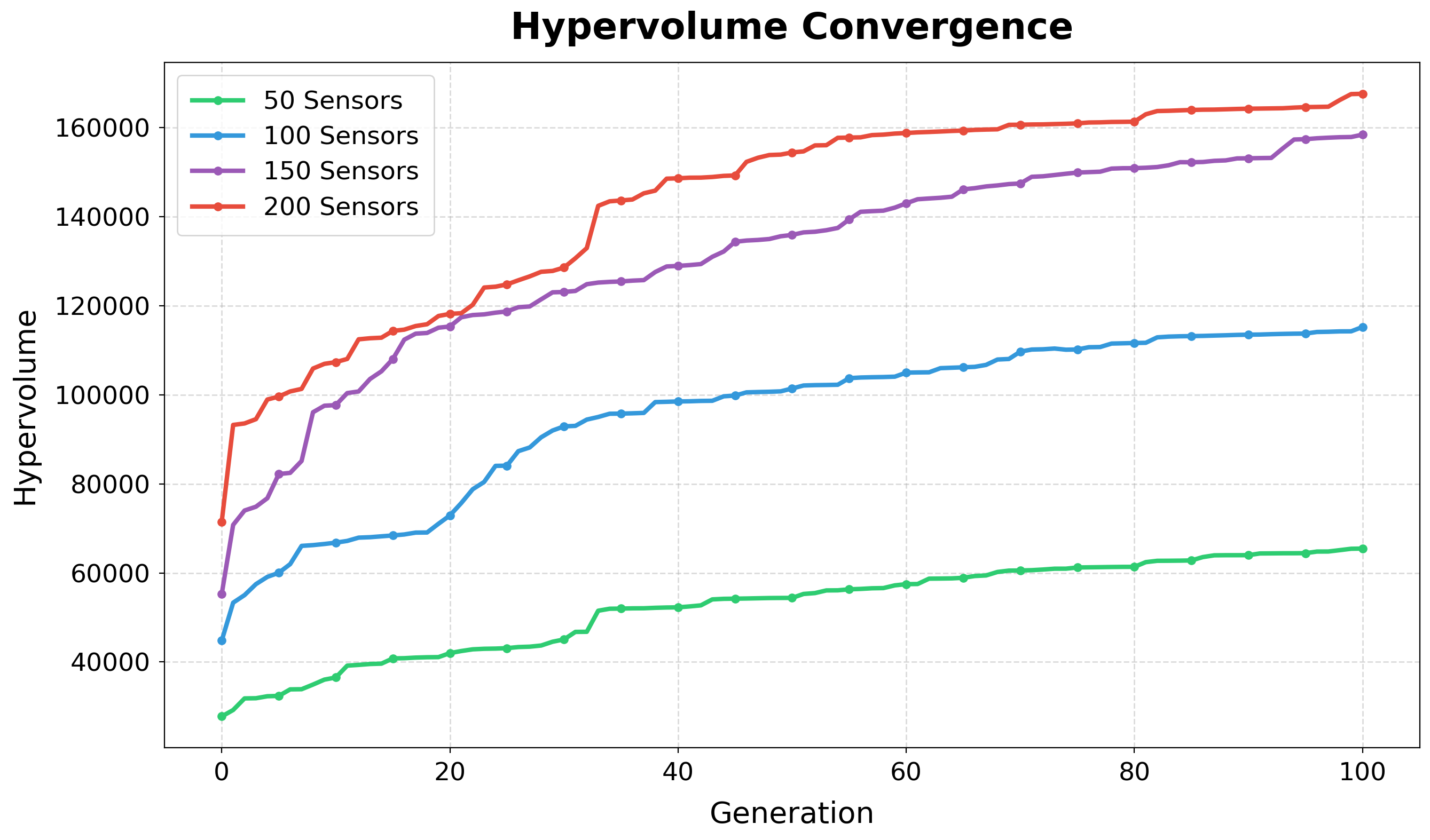}
    \caption{Hypervolume evolution over generations for different sensor densities.}
    \label{fig:hv_results}
\end{figure}

We use hypervolume (HV) as the primary indicator because it jointly reflects convergence and diversity by measuring the dominated area in the bi-objective space. Fig.~\ref{fig:hv_results} shows that HV increases monotonically across all four topologies and then gradually stabilizes, indicating that OG-MOEA/D consistently improves the quality of the non-dominated set over time. {Because exposure is kept in its raw physical scale, absolute HV magnitudes across different sensor-network sizes are descriptive of different objective scales and are not used for cross-topology algorithmic ranking.}

A first notable observation from Fig.~\ref{fig:hv_results} is the clear ordering of the final HV values with respect to sensor density: the 200-sensor case achieves the largest dominated region, followed by 150, 100, and 50 sensors. This trend is expected, since denser deployments provide richer overlapping sensing fields and therefore allow the algorithm to discover paths with substantially higher achievable exposure. At the end of the run, the 200- and 150-sensor topologies reach markedly higher HV values than the 100- and 50-sensor settings, confirming that the proposed algorithm effectively exploits the additional sensing opportunities available in denser networks.

The shape of the HV curves also reveals distinct convergence characteristics. In the 50-sensor case, HV grows relatively slowly and exhibits several visible jumps, suggesting that improvements mainly occur when the algorithm discovers new feasible corridors with noticeably higher exposure. The 100-sensor curve displays a similar staircase pattern, but with stronger mid-run gains, particularly around the first third of the run. In contrast, the 150-sensor topology shows smoother and more regular progress, indicating that once reasonably good path structures are found, the algorithm mainly performs incremental refinement. The 200-sensor case achieves the highest HV throughout most of the run and still exhibits noticeable gains in the later generations, which suggests that although the search space is more complex, the richer sensing field offers more room for improvement.

Another important feature is that none of the curves become completely flat too early. Even in later generations, all four topologies continue to improve, albeit more slowly. This behavior indicates that OG-MOEA/D maintains diversity well enough to avoid premature collapse to a narrow portion of the front. Overall, Fig.~\ref{fig:hv_results} supports the conclusion that the proposed method exhibits reliable convergence and sustained exploration under different sensing densities.

\subsubsection{Evaluating convergence and diversity through Pareto front analysis}
\begin{figure*}[t]
    \centering
    \begin{subfigure}[b]{0.42\textwidth}
        \includegraphics[width=\textwidth]{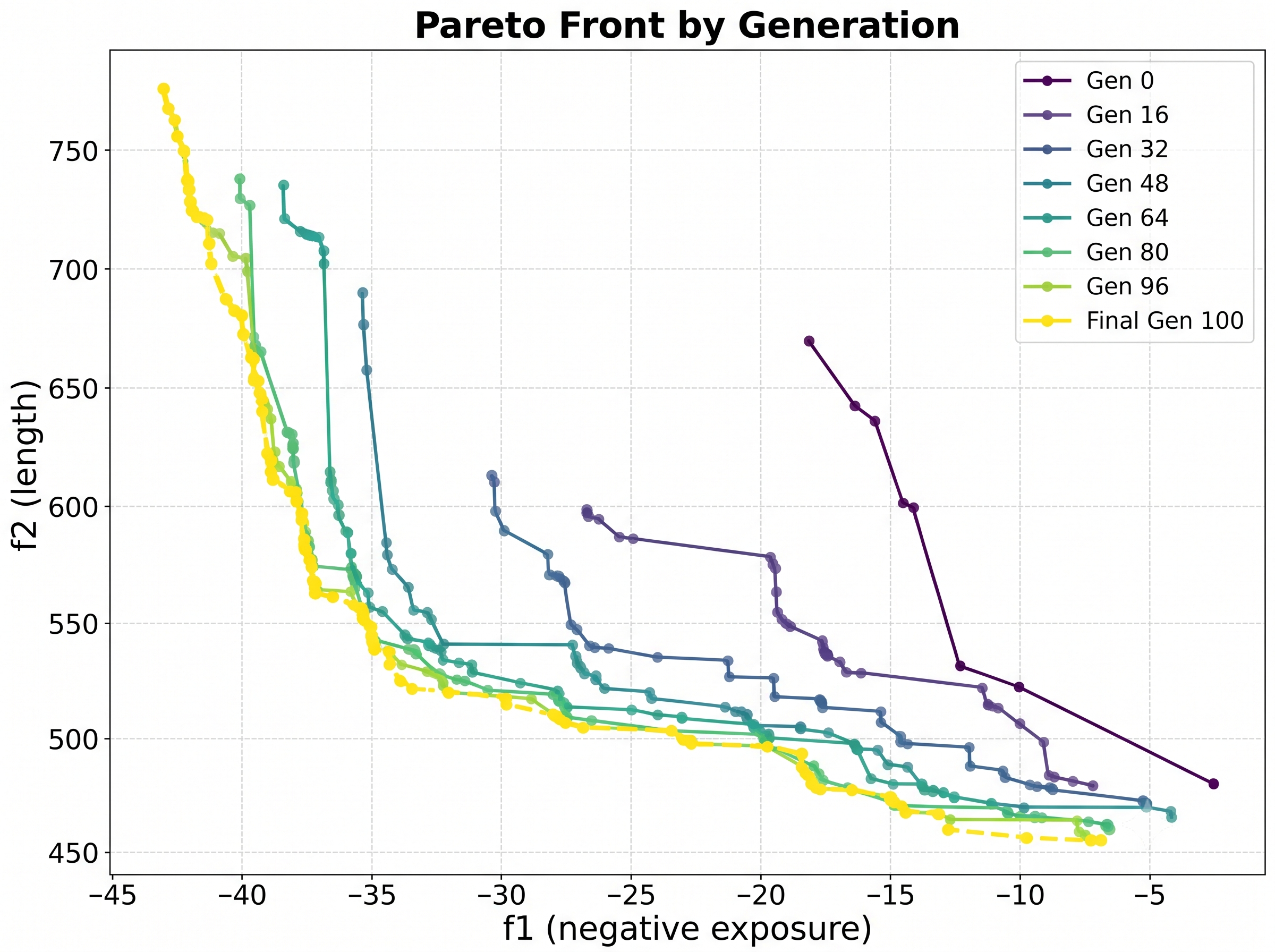}
        \caption{50 Sensors}
        \label{fig:pf_50}
    \end{subfigure}
    \hfill
    \begin{subfigure}[b]{0.42\textwidth}
        \includegraphics[width=\textwidth]{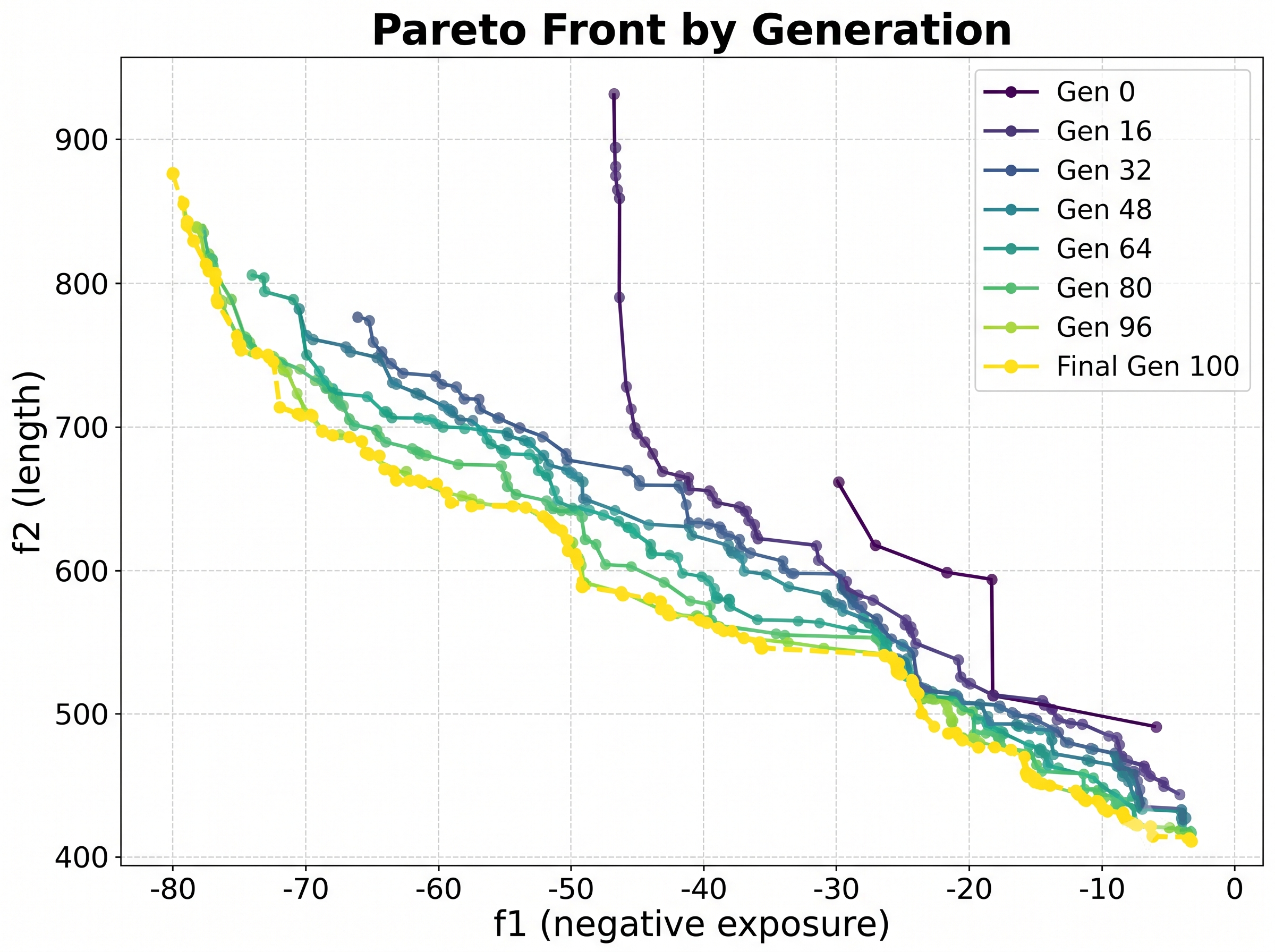}
        \caption{100 Sensors}
        \label{fig:pf_100}
    \end{subfigure}

    \vspace{0.5cm}

    \begin{subfigure}[b]{0.42\textwidth}
        \includegraphics[width=\textwidth]{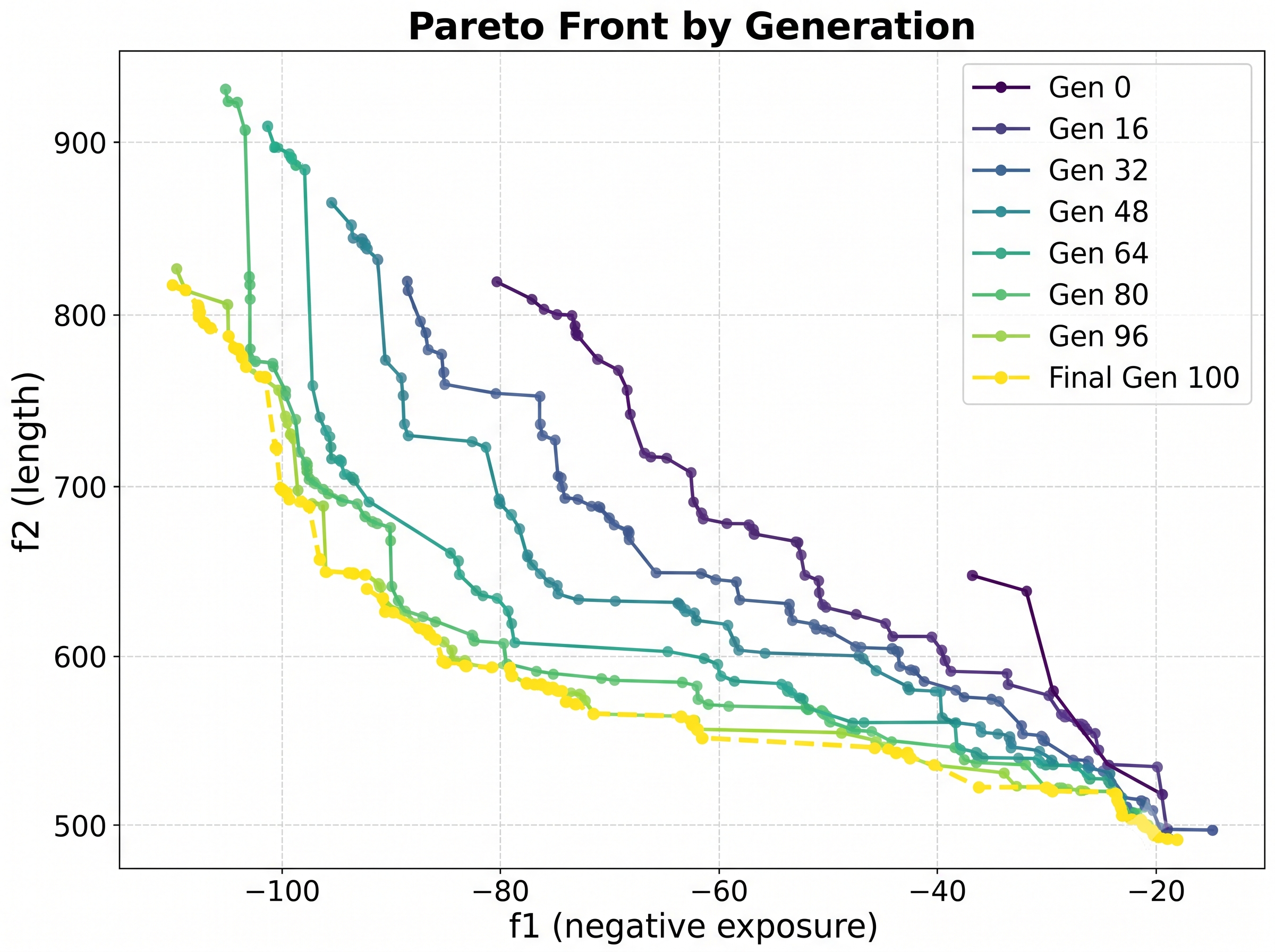}
        \caption{150 Sensors}
        \label{fig:pf_150}
    \end{subfigure}
    \hfill
    \begin{subfigure}[b]{0.42\textwidth}
        \includegraphics[width=\textwidth]{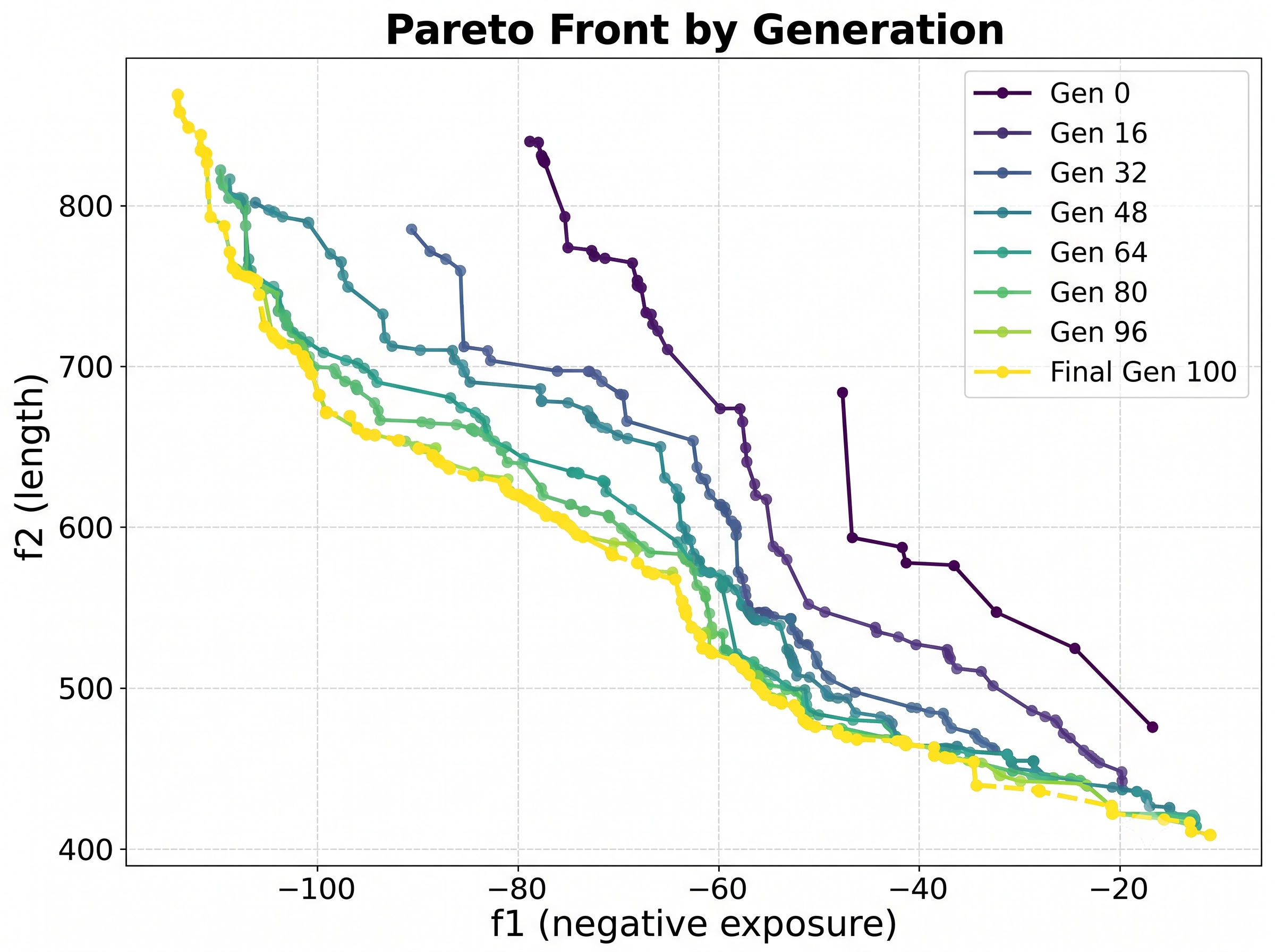}
        \caption{200 Sensors}
        \label{fig:pf_200}
    \end{subfigure}

    \caption{Pareto Front evolution over generations for different sensor densities.}
    \label{fig:pareto_results}
\end{figure*}

Fig.~\ref{fig:pareto_results} complements the HV analysis by showing how the Pareto fronts evolve across generations. In all four topologies, the fronts move progressively toward the lower-left region of the objective space, corresponding to shorter paths and more negative $f_1$ values, i.e., higher accumulated exposure. This consistent displacement confirms that OG-MOEA/D improves both objectives simultaneously as the search proceeds.
The 50-sensor topology exhibits the narrowest objective span, which is consistent with the relatively limited sensing redundancy available in sparse deployments. Here, most of the gain is achieved by gradually smoothing the front and pushing it leftward in moderate steps. In the 100-sensor case, the front evolution is more structured: the initial front contains clearly suboptimal segments, including a steep near-vertical portion, whereas later generations replace these irregular shapes with a smoother and more coherent trade-off curve. This indicates that the algorithm not only improves objective values but also progressively removes unstable path families.

The 150- and 200-sensor cases show a much broader horizontal span, meaning that denser networks allow substantially larger exposure values to be achieved. In both cases, later fronts extend farther to the left while simultaneously dropping downward, which indicates that OG-MOEA/D continues to refine both exposure and path length rather than improving only one objective. Moreover, the final fronts in these denser settings display clearer knee regions. This is practically meaningful because it suggests the existence of solutions where a modest increase in path length yields a comparatively large gain in exposure, offering attractive trade-offs for decision-makers.
Overall, Fig.~\ref{fig:pareto_results} shows that OG-MOEA/D does not merely return isolated non-dominated points; instead, it progressively constructs smooth and well-distributed trade-off fronts across all four topologies. This observation is fully consistent with the HV trends discussed above.

\begin{figure*}[t]
    \centering
    \setlength{\tabcolsep}{4pt}
    \renewcommand{\arraystretch}{1.1}
    \begin{tabular}{cc}
        \multicolumn{1}{c}{\small \textbf{Gen 25}} &
        \multicolumn{1}{c}{\small \textbf{Gen 50}} \\

        \begin{minipage}[t]{0.47\textwidth}
            \centering
            \small \textbf{50 Sensors}\\
            \includegraphics[width=\linewidth]{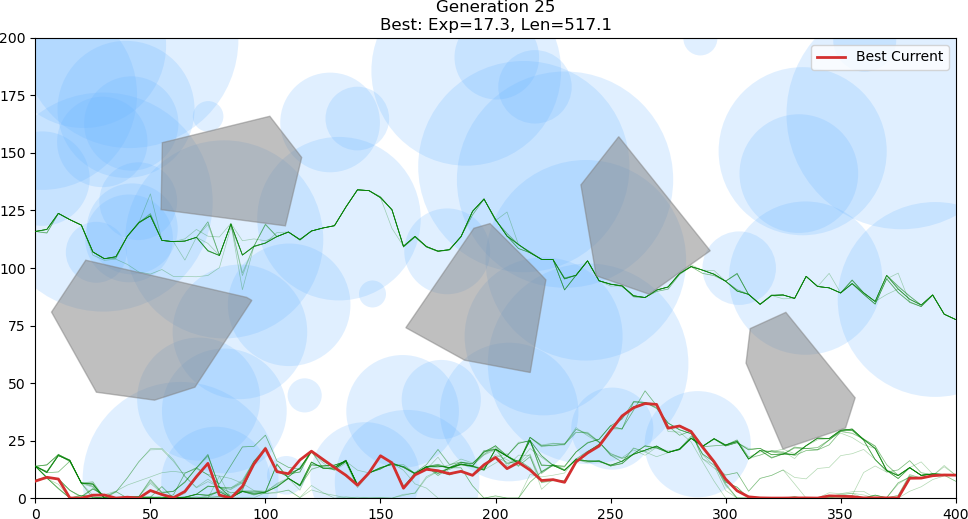}
        \end{minipage}
        &
        \begin{minipage}[t]{0.47\textwidth}
            \centering
            \small \textbf{50 Sensors}\\
            \includegraphics[width=\linewidth]{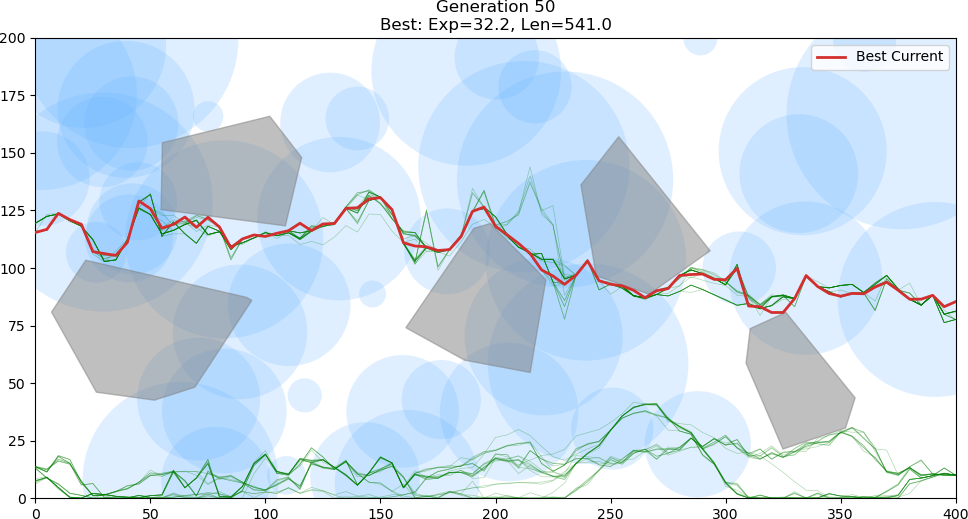}
        \end{minipage}
        \\

        \begin{minipage}[t]{0.47\textwidth}
            \centering
            \small \textbf{100 Sensors}\\
            \includegraphics[width=\linewidth]{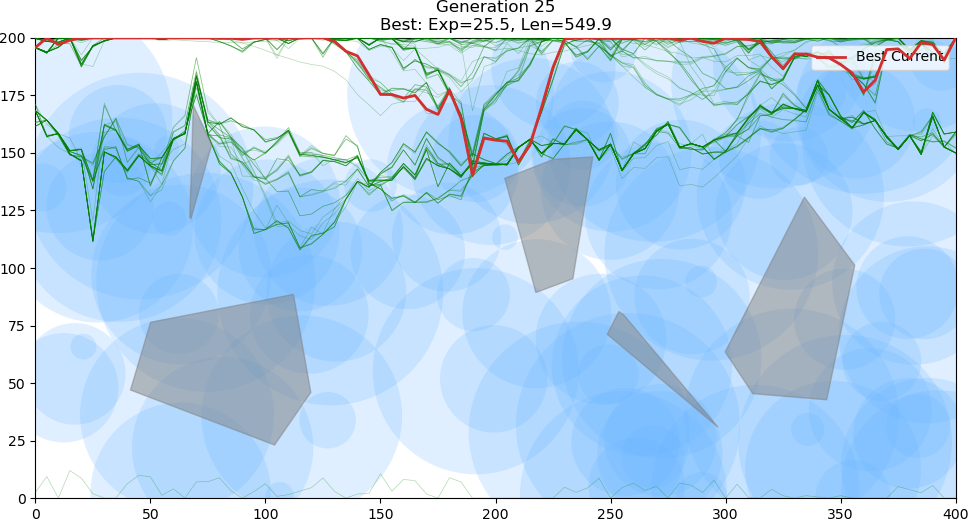}
        \end{minipage}
        &
        \begin{minipage}[t]{0.47\textwidth}
            \centering
            \small \textbf{100 Sensors}\\
            \includegraphics[width=\linewidth]{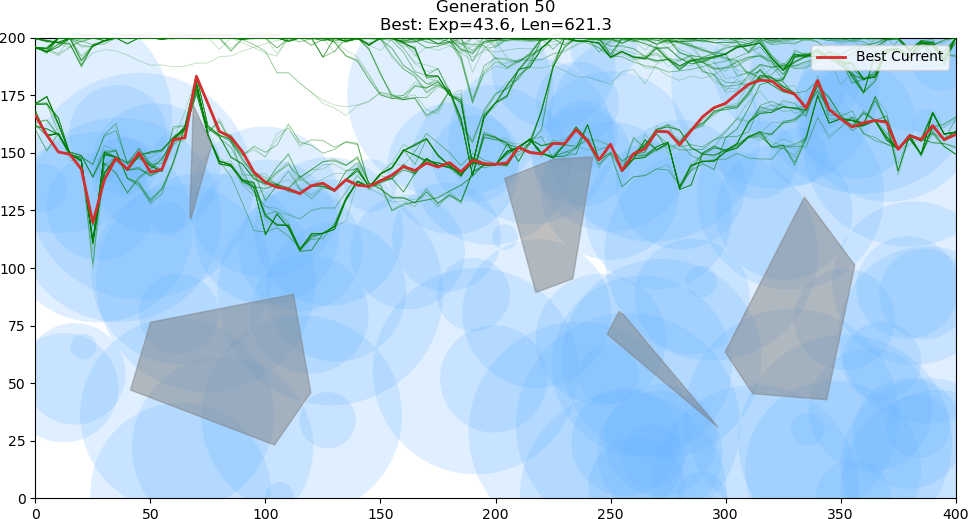}
        \end{minipage}
        \\

        \begin{minipage}[t]{0.47\textwidth}
            \centering
            \small \textbf{150 Sensors}\\
            \includegraphics[width=\linewidth]{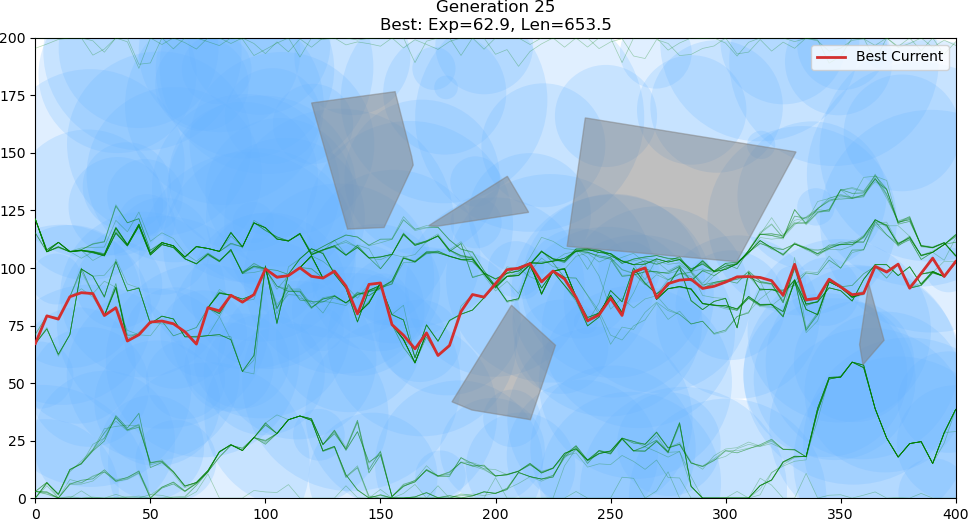}
        \end{minipage}
        &
        \begin{minipage}[t]{0.47\textwidth}
            \centering
            \small \textbf{150 Sensors}\\
            \includegraphics[width=\linewidth]{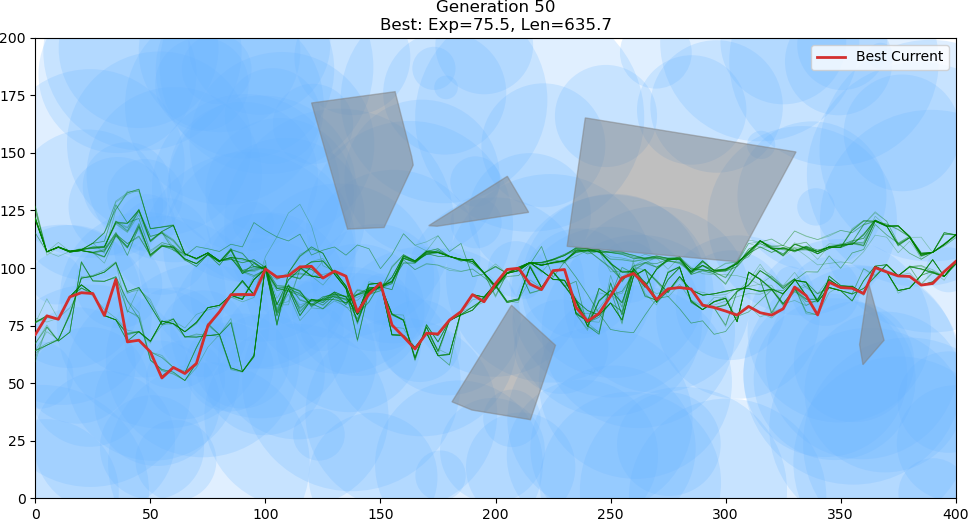}
        \end{minipage}
        \\

        \begin{minipage}[t]{0.47\textwidth}
            \centering
            \small \textbf{200 Sensors}\\
            \includegraphics[width=\linewidth]{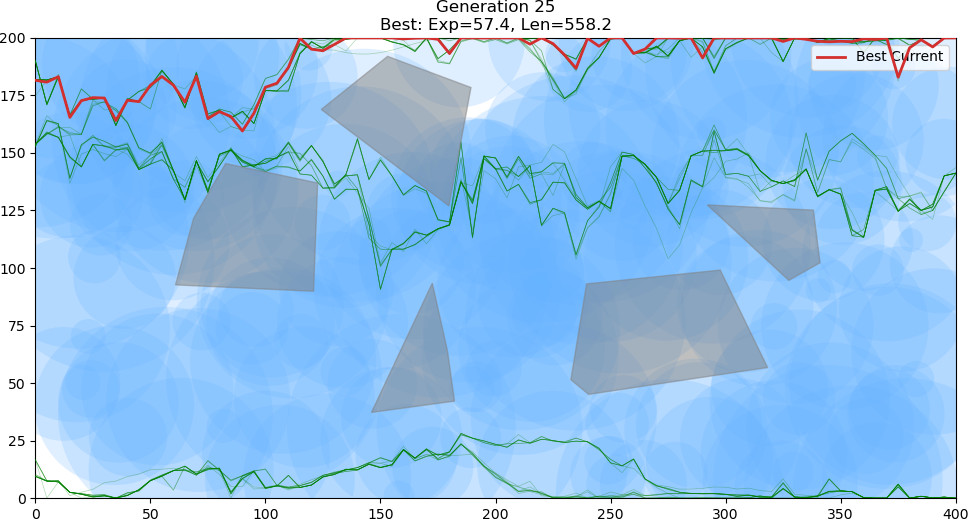}
        \end{minipage}
        &
        \begin{minipage}[t]{0.47\textwidth}
            \centering
            \small \textbf{200 Sensors}\\
            \includegraphics[width=\linewidth]{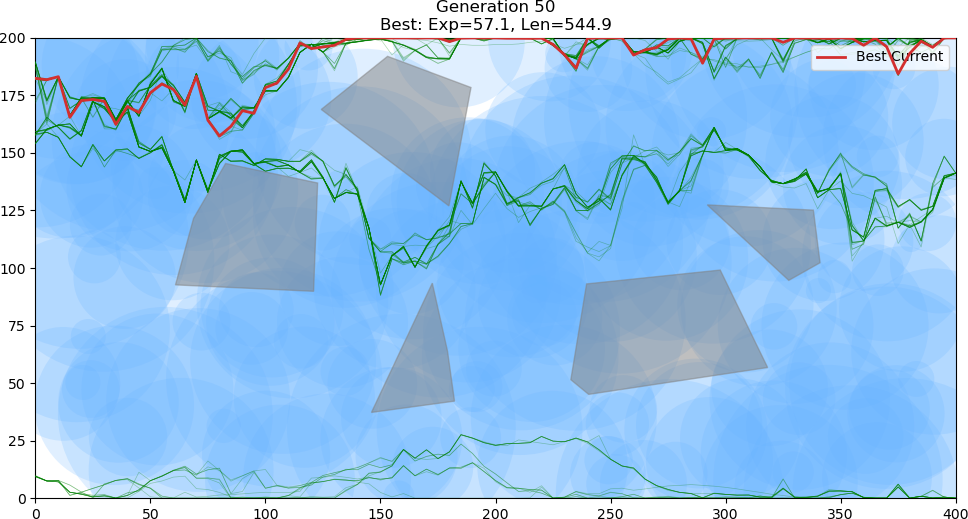}
        \end{minipage}
        \\
    \end{tabular}
    \caption{Representative visual evolution of the discovered paths in the early and middle stages of optimization across different sensor densities.}
    \label{fig:evolution_grid_1}
\end{figure*}
\clearpage
\begin{figure*}[t]
    \centering
    \setlength{\tabcolsep}{4pt}
    \renewcommand{\arraystretch}{1.1}
    \begin{tabular}{cc}
        \multicolumn{1}{c}{\small \textbf{Gen 75}} &
        \multicolumn{1}{c}{\small \textbf{Gen 100}} \\

        \begin{minipage}[t]{0.47\textwidth}
            \centering
            \small \textbf{50 Sensors}\\
            \includegraphics[width=\linewidth]{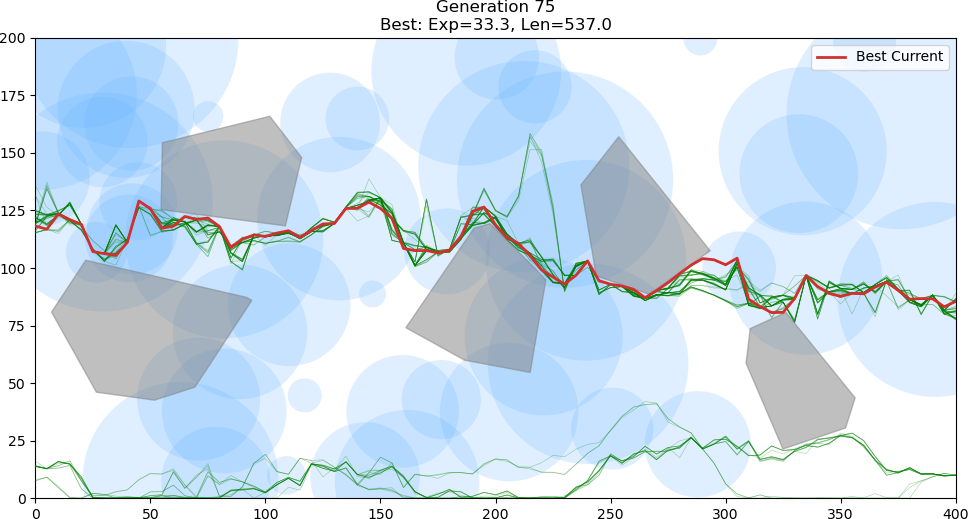}
        \end{minipage}
        &
        \begin{minipage}[t]{0.47\textwidth}
            \centering
            \small \textbf{50 Sensors}\\
            \includegraphics[width=\linewidth]{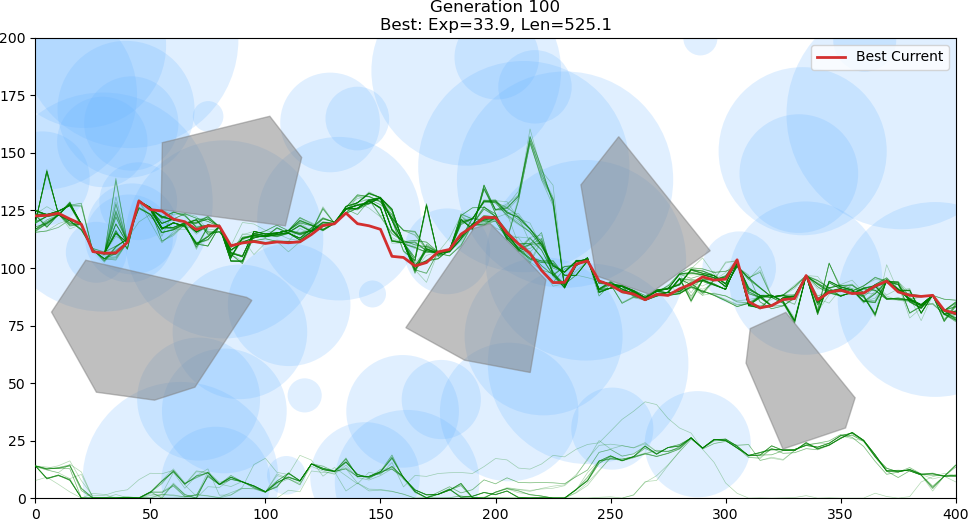}
        \end{minipage}
        \\

        \begin{minipage}[t]{0.47\textwidth}
            \centering
            \small \textbf{100 Sensors}\\
            \includegraphics[width=\linewidth]{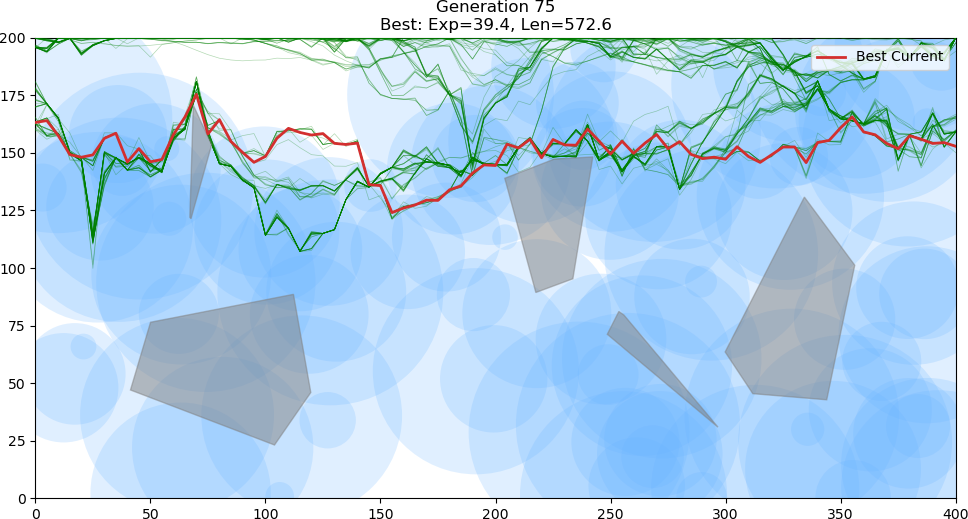}
        \end{minipage}
        &
        \begin{minipage}[t]{0.47\textwidth}
            \centering
            \small \textbf{100 Sensors}\\
            \includegraphics[width=\linewidth]{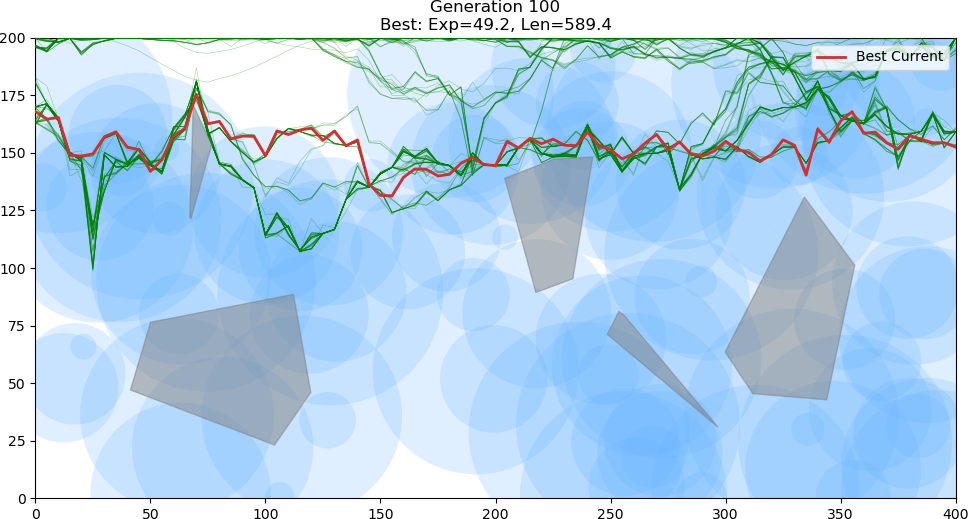}
        \end{minipage}
        \\

        \begin{minipage}[t]{0.47\textwidth}
            \centering
            \small \textbf{150 Sensors}\\
            \includegraphics[width=\linewidth]{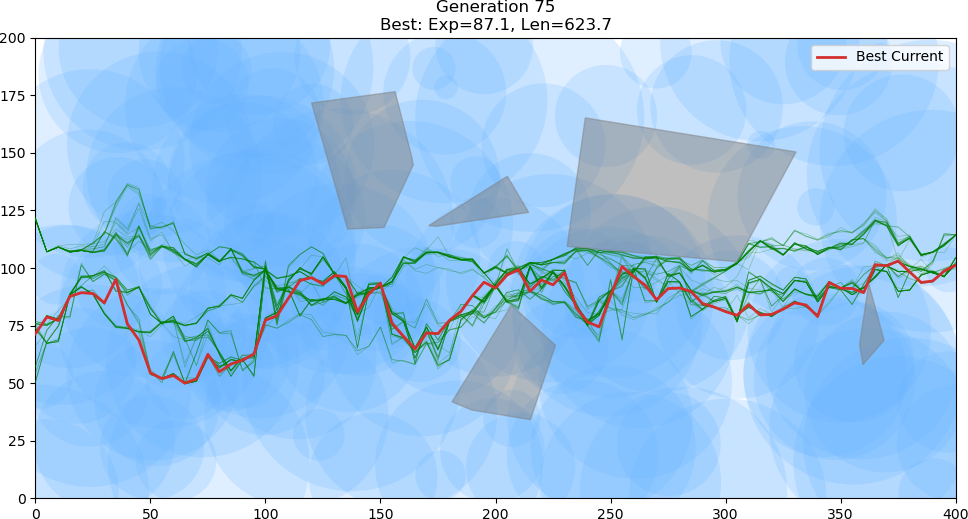}
        \end{minipage}
        &
        \begin{minipage}[t]{0.47\textwidth}
            \centering
            \small \textbf{150 Sensors}\\
            \includegraphics[width=\linewidth]{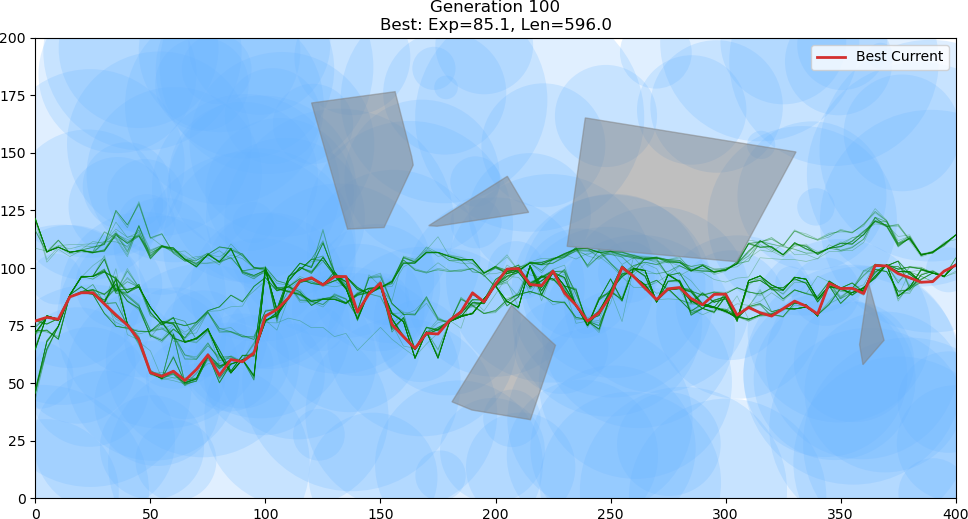}
        \end{minipage}
        \\

        \begin{minipage}[t]{0.47\textwidth}
            \centering
            \small \textbf{200 Sensors}\\
            \includegraphics[width=\linewidth]{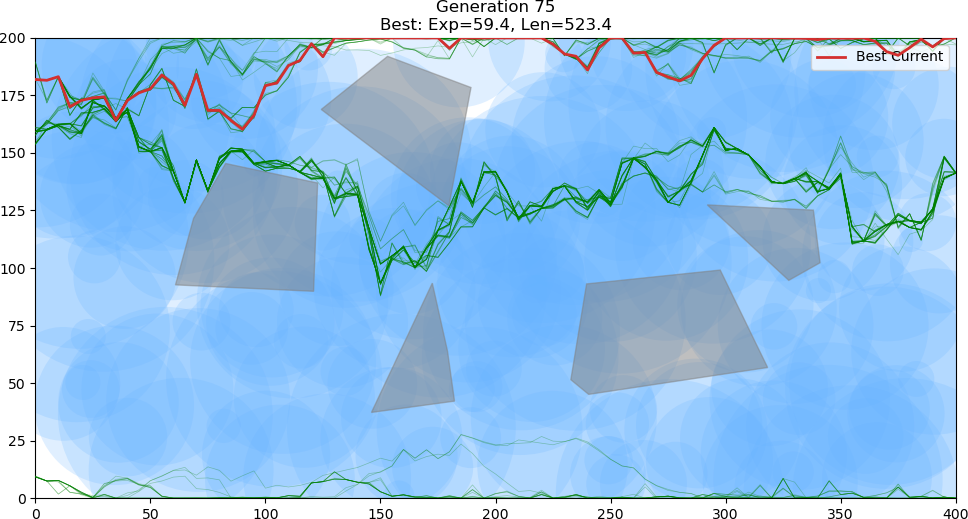}
        \end{minipage}
        &
        \begin{minipage}[t]{0.47\textwidth}
            \centering
            \small \textbf{200 Sensors}\\
            \includegraphics[width=\linewidth]{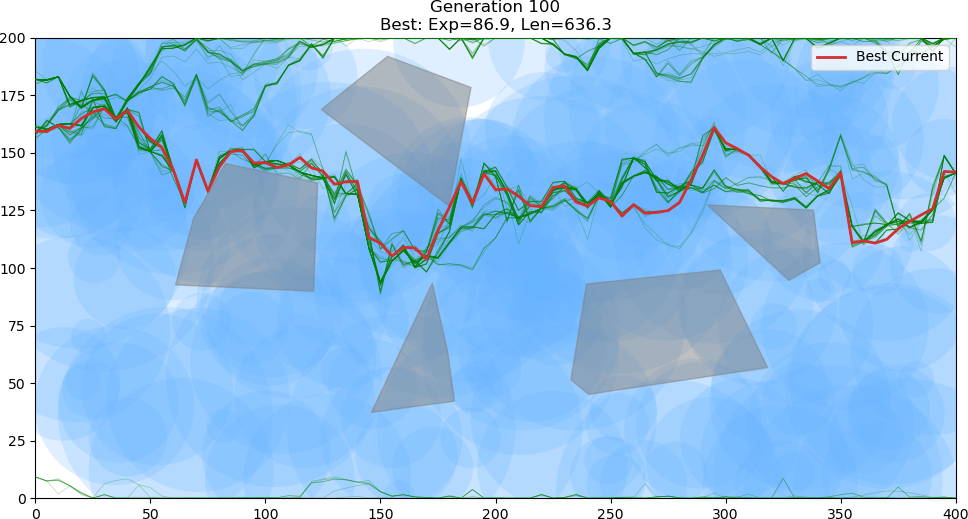}
        \end{minipage}
        \\
    \end{tabular}
    \caption{Representative visual evolution of the discovered paths in the late stage of optimization across different sensor densities.}
    \label{fig:evolution_grid_2}
\end{figure*}
\clearpage

\subsubsection{Qualitative evolution of discovered paths}
The path snapshots in Figs.~\ref{fig:evolution_grid_1} and~\ref{fig:evolution_grid_2} provide a qualitative view of how the search progresses in the spatial domain. In the 50-sensor case, the improvement from Gen~25 to Gen~100 is visually pronounced: the path evolves from a relatively irregular route into a more structured traversal that better follows stronger sensing regions while avoiding large unnecessary deviations. This agrees with the staircase-like HV improvements observed in Fig.~\ref{fig:hv_results}, where gains appear to be associated with the discovery of better corridor structures.

For the 100-sensor topology, the path already lies near a high-exposure band in early generations, but it becomes progressively smoother and more stable over time. In the 150- and 200-sensor cases, the qualitative changes are more subtle: the algorithm rapidly identifies traversals that remain within strongly sensed regions and then mainly performs local refinements. These refinements appear as reduced oscillations, better alignment with dense sensing areas, and elimination of redundant bends. Thus, the path evolution figure supports the interpretation that OG-MOEA/D first identifies promising corridors and then exploits them more precisely in later generations.

Another visible trend is that denser networks tend to produce paths that remain close to high-intensity sensing bands across a larger portion of the ROI. This is especially evident in the 150- and 200-sensor cases, where the best paths increasingly stabilize around well-covered corridors. {Therefore, the qualitative behavior in Figs.~\ref{fig:evolution_grid_1} and~\ref{fig:evolution_grid_2} is consistent with the convergence trends observed in objective space and illustrates how later-generation solutions become more structured spatially.}

\subsection{Comparative analysis}
To further assess search capability and computational efficiency, we compare OG-MOEA/D with MIGA, a dedicated single-objective genetic baseline that maximizes exposure only \cite{binh2025solving}. To isolate the optimization behavior from obstacle-induced feasibility effects, this comparison is conducted in an obstacle-free environment. Both methods use the same population size and number of generations. For OG-MOEA/D, we track the extreme subproblem associated with the weight vector $\alpha=(1.0,0.0)$, which corresponds to the exposure-oriented end of the decomposition. {Both methods use population size 100 and 100 generations, i.e., 10,000 offspring-generation opportunities plus initialization, and are measured end-to-end on the same Python/CPU environment. This comparison is retained only as an auxiliary single-objective reference; the main multiobjective evidence is the obstacle-constrained comparison in Section 5.4.}

\subsubsection{Search capability and local optima avoidance}
Table~\ref{tab:ga_comparison} shows that OG-MOEA/D achieves the best exposure in two out of the four tested topologies and remains highly competitive in the other two. The clearest improvement appears in the 50-sensor case, where OG-MOEA/D increases the best exposure from 70.69 to 73.97, corresponding to a gain of approximately 4.64\%. A second clear advantage appears in the 200-sensor setting, where the best exposure improves from 169.21 to 173.28, i.e., about 2.41\%. These are the two cases where improved exploration appears most beneficial: the sparse 50-sensor environment provides fewer good corridors, while the 200-sensor environment creates a more rugged search landscape with many competing alternatives.

At the same time, Table~\ref{tab:ga_comparison} also shows that the superiority of OG-MOEA/D in exposure is not uniform across all cases. In the 100-sensor topology, MIGA attains a slightly higher best exposure (116.35 versus 116.12), while in the 150-sensor topology the difference is extremely small, with OG-MOEA/D improving exposure only marginally (171.91 versus 171.83). These results suggest that the main advantage of OG-MOEA/D is not simply a guaranteed increase in best exposure for every topology, but rather a more robust search behavior across environments with different structural characteristics.

From an algorithmic viewpoint, the gains observed in the 50- and 200-sensor cases are consistent with the neighborhood-based information sharing of MOEA/D. Even when the selected subproblem emphasizes exposure, the solution can still benefit from nearby trade-off directions, which helps preserve exploratory capability and reduces the risk of stagnation. In contrast, the single-objective GA may perform competitively in intermediate cases but appears more vulnerable when the search either becomes too sparse or too structurally complex.

\subsubsection{Computational efficiency}

\begin{table}[t]
    \centering
    \caption{{Comparison of solution quality (exposure) and computational cost (runtime) between OG-MOEA/D and MIGA.}}
    \label{tab:ga_comparison}
    \setlength{\tabcolsep}{6pt}

    \begin{tabular}{lccc}
        \toprule
        \textbf{Topology} & \textbf{Algorithm} & \textbf{Best Exposure} & \textbf{Runtime (s)} \\
        \midrule

        \multirow{2}{*}{\textbf{50 Sensors}} & MIGA & 70.69 & 1016.2 \\
         & \textbf{OG-MOEA/D} & \textbf{73.97} & \textbf{663.0} \\
        \midrule

        \multirow{2}{*}{\textbf{100 Sensors}} & \textbf{MIGA} & \textbf{116.35} & 2694.2 \\
         & OG-MOEA/D & 116.12 & \textbf{2283.0} \\
        \midrule

        \multirow{2}{*}{\textbf{150 Sensors}} & MIGA & 171.83 & 3798.8 \\
         & \textbf{OG-MOEA/D} & \textbf{171.91} & \textbf{2758.9} \\
        \midrule

        \multirow{2}{*}{\textbf{200 Sensors}} & MIGA & 169.21 & 4676.6 \\
         & \textbf{OG-MOEA/D} & \textbf{173.28} & \textbf{4390.1} \\
        \bottomrule
    \end{tabular}
    \vspace{-.2cm}
\end{table}

In this auxiliary single-objective comparison, the measured runtime of OG-MOEA/D is lower than that of MIGA in all four tested cases. The observed reductions are approximately 34.7\%, 15.3\%, 27.4\%, and 6.1\% for the 50-, 100-, 150-, and 200-sensor cases, respectively. Because Table~\ref{tab:ga_comparison} is retained as a descriptive auxiliary comparison rather than the main stochastic benchmark, we do not interpret these measurements as a statistically established runtime advantage.
Taken together, Table~\ref{tab:ga_comparison} shows that the exposure-oriented extreme solution of OG-MOEA/D is competitive with MIGA in the tested obstacle-free cases while exhibiting lower measured wall-clock time in these runs. The principal evidence for the proposed multiobjective method is therefore the repeated obstacle-constrained comparison in Section~\ref{sec:comparison2}, rather than this auxiliary MIGA experiment.

{\subsection{Comparison with Baseline Algorithms}}
\label{sec:comparison2}
{
To evaluate the multiobjective performance of OG-MOEA/D, we compare it with two representative multiobjective evolutionary algorithms, MOEA/D and NSGA-II~\cite{zhang2007moead,deb2002nsga2}, under the same obstacle-constrained OE-MEP evaluation protocol. Each algorithm is executed independently for 20 runs on the 50-, 100-, 150-, and 200-sensor topologies, using the same seed index set and the matched population/generation budget specified in Section~5.1.

We use Hypervolume (HV) and Inverted Generational Distance (IGD). Both objectives are minimized, with $f_1=-\mathsf{Exp}$ and $f_2=L$. HV is computed directly in the raw objective space using the fixed reference point
\[
\mathbf{r}_{\rm HV}=(1,L_{\max})=(1,2000),
\]
which is worse than the feasible approximation sets because feasible exposure values satisfy $f_1\le 0$ and path length satisfies $f_2\le L_{\max}$. Because the raw exposure scale changes with sensor density, HV is used for comparing algorithms \emph{within the same topology}; absolute HV values across different topologies are not interpreted as directly comparable algorithmic-quality scores.

For IGD, the exact Pareto front is unknown. For each topology, we therefore construct an empirical reference front from the nondominated union of all feasible solutions produced by all three algorithms across all 20 runs. Both objective dimensions are min--max normalized using bounds from this common empirical set before IGD is calculated. Table~\ref{tab:performance_comparison} reports mean $\pm$ standard deviation. Pairwise differences relative to OG-MOEA/D are assessed with the two-sided Wilcoxon rank-sum (Mann--Whitney $U$) test at $\alpha=0.05$, treating the 20 run-level metric values from each algorithm as independent samples; ``+'' indicates that OG-MOEA/D is significantly better than the corresponding baseline and ``$\approx$'' indicates that the difference is not statistically significant.
}

\begin{table}[t]
    \centering
    \caption{{Multiobjective performance across obstacle-constrained sensor-network topologies over 20 independent runs.}}
    \label{tab:performance_comparison}
    \setlength{\tabcolsep}{5pt}
    \footnotesize
    \begin{tabular}{llccc}
        \toprule
        \textbf{Topology} & \textbf{Metric} & \textbf{OG-MOEA/D} & \textbf{MOEA/D} & \textbf{NSGA-II} \\
        \midrule
        \multirow{2}{*}{\textbf{50 Sensors}}
        & HV  & \textbf{85090.4 $\pm$ 7898.8} & 53663.5 $\pm$ 5991.3 (+) & 72818.9 $\pm$ 2117.8 (+) \\
        & IGD & \textbf{0.141 $\pm$ 0.037} & 0.339 $\pm$ 0.018 (+) & 0.314 $\pm$ 0.018 (+) \\
        \midrule
        \multirow{2}{*}{\textbf{100 Sensors}}
        & HV  & \textbf{139134.7 $\pm$ 13622.6} & 91111.4 $\pm$ 7838.6 (+) & 107386.0 $\pm$ 5350.7 (+) \\
        & IGD & \textbf{0.102 $\pm$ 0.032} & 0.278 $\pm$ 0.031 (+) & 0.259 $\pm$ 0.018 (+) \\
        \midrule
        \multirow{2}{*}{\textbf{150 Sensors}}
        & HV  & \textbf{187614.6 $\pm$ 11643.7} & 134939.9 $\pm$ 7871.5 (+) & 157929.9 $\pm$ 3830.8 (+) \\
        & IGD & \textbf{0.085 $\pm$ 0.035} & 0.233 $\pm$ 0.019 (+) & 0.191 $\pm$ 0.012 (+) \\
        \midrule
        \multirow{2}{*}{\textbf{200 Sensors}}
        & HV  & \textbf{217548.9 $\pm$ 7260.6} & 171703.6 $\pm$ 9532.6 (+) & 204887.0 $\pm$ 5735.1 (+) \\
        & IGD & \textbf{0.061 $\pm$ 0.013} & 0.151 $\pm$ 0.014 (+) & 0.113 $\pm$ 0.009 (+) \\
        \bottomrule
        \multicolumn{5}{l}{\scriptsize (+): OG-MOEA/D significantly better at $\alpha=0.05$.}
    \end{tabular}
\end{table}

\begin{figure*}[htbp]
\centering
\includegraphics[width=0.9\textwidth]{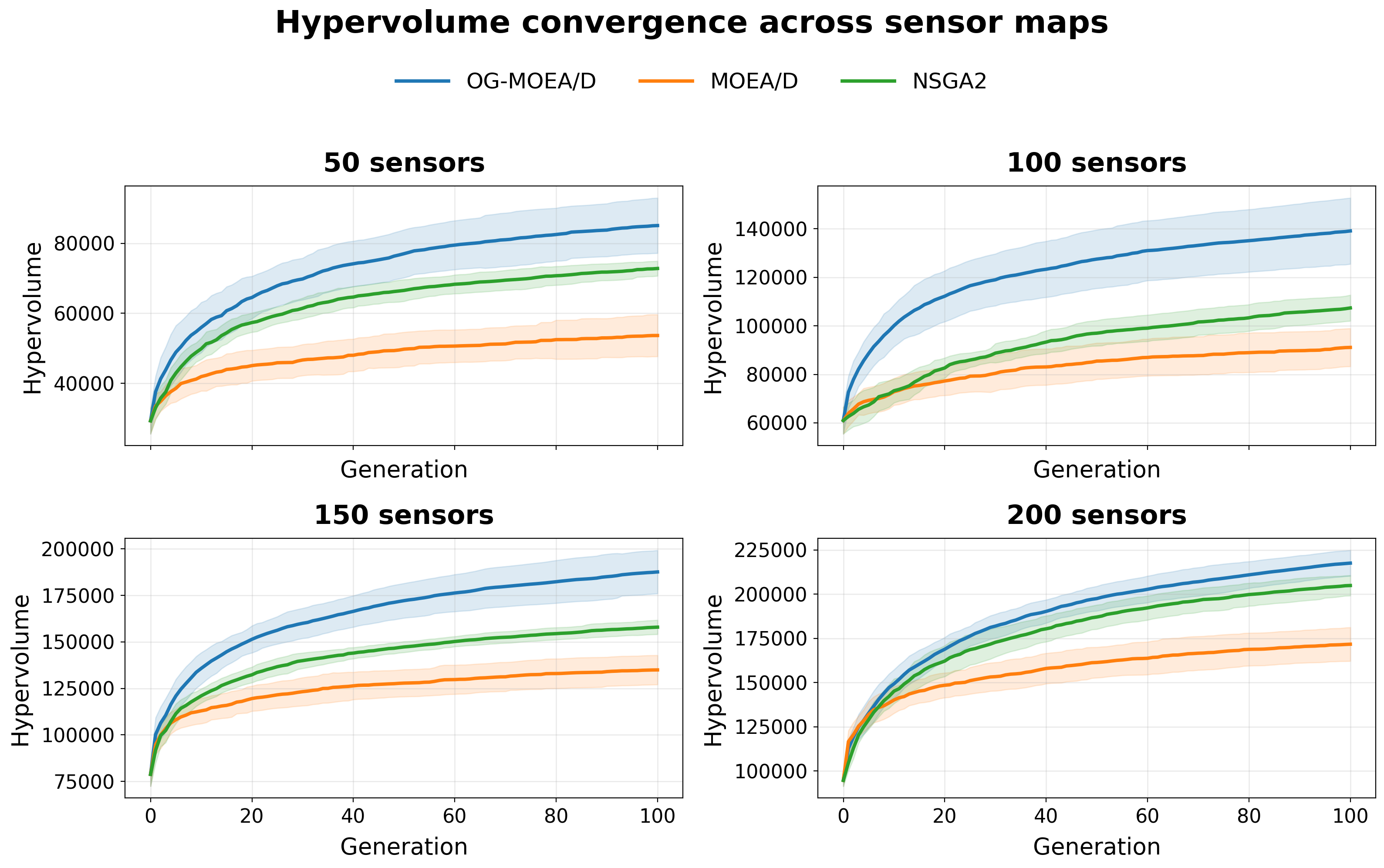}
\caption{{Hypervolume convergence of OG-MOEA/D, MOEA/D, and NSGA-II over 100 generations across four obstacle-constrained sensor-network configurations.}}
\label{fig:hv_convergence}
\end{figure*}

{
Table~\ref{tab:performance_comparison} shows that OG-MOEA/D achieves higher mean HV and lower mean IGD than both baselines in all four tested topologies. Figure~\ref{fig:hv_convergence} provides the corresponding convergence view. These results provide empirical evidence that the \emph{complete} problem-specific OG-MOEA/D design---fixed-grid representation together with feasibility-aware handling---is effective relative to MOEA/D under the considered obstacle-constrained instances. They do not, by themselves, isolate the causal contribution of each individual component, and we therefore avoid attributing the entire performance gap to the repair operator alone.

The 20-run standard deviations also provide a direct measure of stochastic variability. We use the word ``significant'' only for comparisons supported by the stated Wilcoxon test. MIGA remains in Section~5.3 as an auxiliary single-objective historical reference and is not used as the principal evidence for multiobjective superiority.
}

\begin{figure*}[htbp]
\centering
\includegraphics[width=1.0\textwidth]{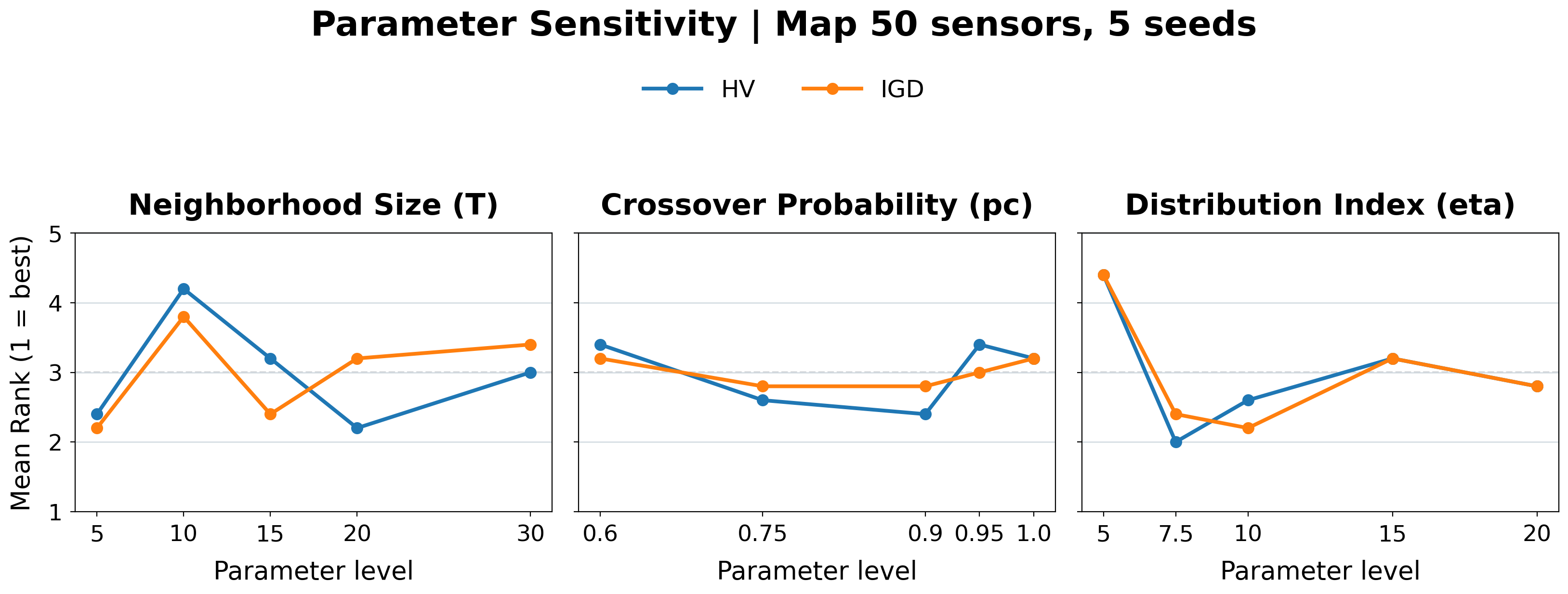}
\caption{{Parameter sensitivity of OG-MOEA/D on HV and IGD using Friedman mean ranks over five independent seeds.}}
\label{fig:sensitivity}
\end{figure*}

{\subsection{Parameter Sensitivity Analysis}}
\label{sec:sensitivity}
{
We further examine the sensitivity of OG-MOEA/D to three influential parameter groups using five independent seeds on the 50-sensor topology: neighborhood size $T\in\{5,10,15,20,30\}$, crossover probability $p_c\in\{0.6,0.75,0.9,0.95,1.0\}$, and a common SBX/polynomial-mutation distribution index $\eta=\eta_c=\eta_m\in\{5,7.5,10,15,20\}$. The non-parametric Friedman mean rank is reported for HV and IGD across the five run replicates for each tested setting; lower mean rank is better. Figure~\ref{fig:sensitivity} summarizes the results.

For neighborhood size, moderate settings are more favorable than the extreme values, with $T=20$ attaining the best HV rank in the tested sweep. For crossover, performance improves as $p_c$ increases toward $0.9$ and becomes less favorable at the most aggressive settings. For the distribution indices, the best region lies around $\eta=7.5$--$10$, and the default $\eta_c=\eta_m=10$ therefore falls within the empirically favorable range. These observations are interpreted as sensitivity trends for the tested setting rather than as evidence of a unique globally optimal parameter configuration.
}

{
\paragraph{Grid-resolution sensitivity.}
Because the fixed representation is $x$-monotone, its resolution directly controls representational capacity. We test $\Delta x\in\{2,5,10,20,40,80\}$, corresponding to $M\in\{200,80,40,20,10,5\}$ intervals and $n=M+1\in\{201,81,41,21,11,6\}$ encoded grid points. Each configuration is executed for five independent runs. The mean $\pm$ standard-deviation HV values are $88145.4\pm9279.2$, $88610.5\pm17101.8$, $85167.7\pm5030.5$, $74389.3\pm9687.8$, $64428.6\pm10030.5$, and $36101.0\pm5646.9$, respectively. Coarsening the grid substantially reduces the quality of the approximation set, whereas increasing the resolution from the default $n=81$ to $n=201$ does not improve mean HV despite increasing the decision dimension. The default $M=80$ ($n=81$, $\Delta x=5$) is therefore retained as a reasonable quality--complexity trade-off for the tested environment.
}

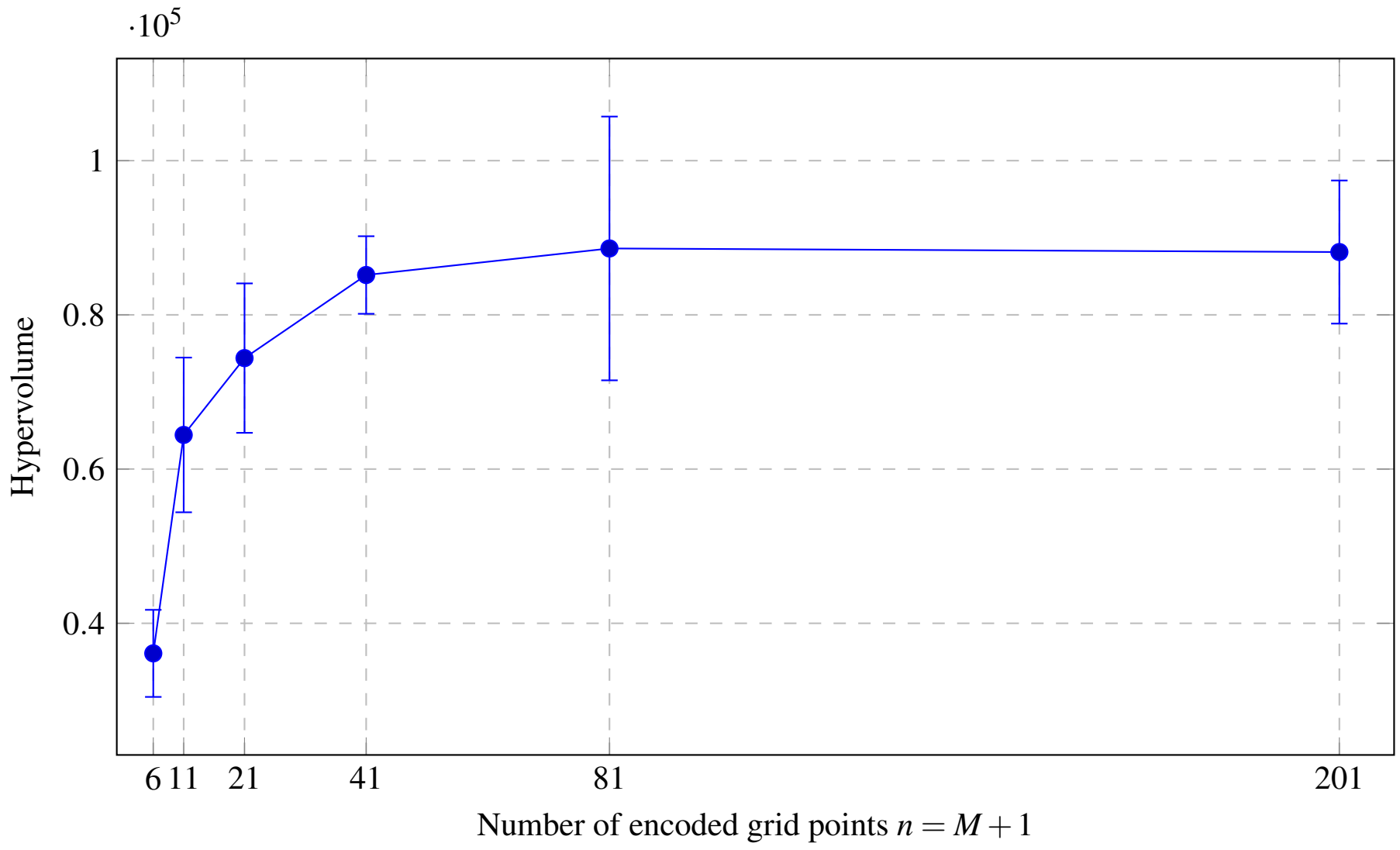
\begin{figure}[htbp]
\centering
\begin{tikzpicture}
\begin{axis}[
    width=0.78\linewidth,
    height=0.47\linewidth,
    xlabel={Number of encoded grid points $n=M+1$},
    ylabel={Hypervolume},
    xmin=0, xmax=210,
    xtick={6,11,21,41,81,201},
    ymajorgrids=true,
    xmajorgrids=true,
    grid style={dashed},
    error bars/y dir=both,
    error bars/y explicit
]
\addplot+[mark=*] coordinates {
    (6,36101.0) +- (0,5646.9)
    (11,64428.6) +- (0,10030.5)
    (21,74389.3) +- (0,9687.8)
    (41,85167.7) +- (0,5030.5)
    (81,88610.5) +- (0,17101.8)
    (201,88145.4) +- (0,9279.2)
};
\end{axis}
\end{tikzpicture}
\caption{{Grid-resolution sensitivity of OG-MOEA/D on the 50-sensor topology. Error bars show one standard deviation over five independent runs. The tested grid-point counts $n=\{201,81,41,21,11,6\}$ correspond to $\Delta x=\{2,5,10,20,40,80\}$ for $W=400$.}}
\label{fig:grid_sensitivity}
\end{figure}

\FloatBarrier
\section{Conclusions}\label{sec:conclusion}

{This paper studied obstacle-aware maximal-exposure path planning as a bi-objective trade-off between accumulated sensing exposure and path length under heterogeneous truncated sensing, LoS occlusion, and polygonal obstacle constraints. The resulting Pareto set should be interpreted as a surveillance-strength/traversal-efficiency trade-off: maximal exposure characterizes strongly sensed crossings and does not model the behavior of a rational evasive intruder. OG-MOEA/D combines a fixed-grid $x$-monotone path representation and explicit feasibility handling with standard MOEA/D search. In the repeated obstacle-constrained comparison, OG-MOEA/D obtains higher mean HV and lower mean IGD than MOEA/D and NSGA-II for all four tested sensor-network sizes. The parameter study identifies empirically favorable regions for the tested settings, while the grid-resolution study shows a clear loss in approximation quality for coarse discretizations and little additional mean-HV benefit when increasing from $n=81$ to $n=201$ grid points. The present evaluation is limited to synthetic two-dimensional environments with convex polygonal obstacles, one explicitly specified heterogeneous sensing regime, and the restricted $x$-monotone path family $\mathcal{P}^{\uparrow}_M$. It should therefore be interpreted as controlled simulation evidence rather than validation of a complete real-world deployment, a systematic characterization of heterogeneity strength, or an unrestricted path optimum. Future work will consider sensitivity to different degrees of sensing heterogeneity, non-monotone/variable-length representations, non-convex and dynamic obstacles, three-dimensional motion, measured deployment data, and vulnerability-oriented minimal-exposure analysis. For the latter, the same sensing and feasibility machinery can be retained while reversing the exposure objective, yielding a trade-off between stealthiness and traversal length.}









\section*{Declarations}
The authors declare no conflict of interest, particularly no financial and personal relationships with other people or organizations that could inappropriately influence (bias) this work.

\section*{Availability of Data and Materials}
The data and implementation source code used in this study are provided with the submission as supplementary materials.
\nocite{*}
\bibliographystyle{ieeetr}
\bibliography{sn-bibliography,revision_references}%

\end{document}